\documentclass[sigconf, screen]{acmart}
\renewcommand\footnotetextcopyrightpermission[1]{}
\usepackage{subfig}
\usepackage{algorithm}      % 提供算法浮动体环境
\usepackage{algpseudocode} 
\usepackage{multicol}
\usepackage{array} 
\usepackage{multirow}
\usepackage{booktabs} 
\usepackage{graphicx}
\usepackage{colortbl}

\begin{document}

%%
%% The "title" command has an optional parameter,
%% allowing the author to define a "short title" to be used in page headers.
\title{VGNC: Reducing the Overfitting of Sparse-view 3DGS via Validation-guided Gaussian Number Control
%ValidationGaussian: Validation-Guided Adaptive Gaussian for Overfitting-Free Sparse-View 3DGS
}

%ValutationGaussian：解决过拟合问题的验证引导的自适应3DGS

%%
%% The "author" command and its associated commands are used to define
%% the authors and their affiliations.
%% Of note is the shared affiliation of the first two authors, and the
%% "authornote" and "authornotemark" commands
%% used to denote shared contribution to the research.

% \author{Lifeng Lin}
% \authornote{Both authors contributed equally to this research.}
% % \email{trovato@corporation.com}
% % \orcid{1234-5678-9012}
% \author{Rongfeng Lu}
% \authornotemark[1]
% \email{rongfeng-lu@hdu.edu.cn}
% \affiliation{%
%   \institution{Hangzhou Dianzi University}
%   % \city{Hangzhou}
%   % \state{Zhejiang}
%   % \country{China}
% }
\author{Lifeng Lin}
\authornote{Both authors contributed equally to this research.}
\email{lifeng_lin@hdu.edu.cn}
\affiliation{%
  \institution{Hangzhou Dianzi University}
  \city{ }
  \state{ }
  \country{ }
}
% \orcid{1234-5678-9012}
\author{Rongfeng Lu}
\authornotemark[1]
\email{rongfeng-lu@hdu.edu.cn}
\affiliation{%
  \institution{Hangzhou Dianzi University}
    \city{ }
  \state{ }
  \country{ }
  % \city{Hangzhou}
  % \state{Zhejiang}
  % \country{China}
}
\author{Quan Chen}
% \email{chenquan@alu.hdu.edu.cn}
\affiliation{%
  \institution{Hangzhou Dianzi University}
  \city{ }
  \state{ }
  \country{ }
  }
\author{Haofan Ren}
\affiliation{%
  \institution{Hangzhou Dianzi University}
  \city{ }
  \state{ }
  \country{ }
  }
% \email{rhfkris@gmail.com}
\author{Ming Lu}
\affiliation{%
  \institution{Intel Labs China}
    \city{ }
  \state{ }
  \country{ }
  % \city{Beijing}
  % \country{China}
  }
% \email{lu199192@gmail.com}
\author{Yaoqi Sun}
\affiliation{%
  \institution{Lishui University}
   \city{ }
  \state{ }
  \country{ }
  % \city{Lishui}
  % \state{Zhejiang}
  % \country{China}
  }
% \email{sunyq2233@163.com}
\author{Chenggang Yan}
\affiliation{%
  \institution{Hangzhou Dianzi University}
  \city{ }
  \state{ }
  \country{ }
  }
% \email{cgyan@hdu.edu.cn}
\author{Anke Xue}
\affiliation{%
  \institution{Hangzhou Dianzi University}
  \city{ }
  \state{ }
  \country{ }
  }
% \email{akxue@hdu.edu.cn}

%%
%% By default, the full list of authors will be used in the page
%% headers. Often, this list is too long, and will overlap
%% other information printed in the page headers. This command allows
%% the author to define a more concise list
%% of authors' names for this purpose.
% \renewcommand{\shortauthors}{et al.}

%%
%% The abstract is a short summary of the work to be presented in the
%% article.
\begin{abstract}
Sparse-view 3D reconstruction is a fundamental yet challenging task in practical 3D reconstruction applications. Recently, many methods based on the 3D Gaussian Splatting (3DGS) framework have been proposed to address sparse-view 3D reconstruction. Although these methods have made considerable advancements, they still show significant issues with overfitting. To reduce the overfitting, we introduce VGNC, a novel Validation-guided Gaussian Number Control (VGNC) approach based on generative novel view synthesis (NVS) models. To the best of our knowledge, this is the first attempt to alleviate the overfitting issue of sparse-view 3DGS with generative validation images. Specifically, we first introduce a validation image generation method based on a generative NVS model. We then propose a Gaussian number control strategy that utilizes generated validation images to determine the optimal Gaussian numbers, thereby reducing the issue of overfitting. We conducted detailed experiments on various sparse-view 3DGS baselines and datasets to evaluate the effectiveness of VGNC. Extensive experiments show that our approach not only reduces overfitting but also improves rendering quality on the test set while decreasing the number of Gaussian points. This reduction lowers storage demands and accelerates both training and rendering. The code will be released.
\end{abstract}

%%
%% The code below is generated by the tool at http://dl.acm.org/ccs.cfm.
%% Please copy and paste the code instead of the example below.
%%


%%
%% Keywords. The author(s) should pick words that accurately describe
%% the work being presented. Separate the keywords with commas.
%% A "teaser" image appears between the author and affiliation
%% information and the body of the document, and typically spans the
%% page.
\begin{teaserfigure}
    \centering
  \subfloat[Train views]{
  \includegraphics[width=0.33\textwidth]{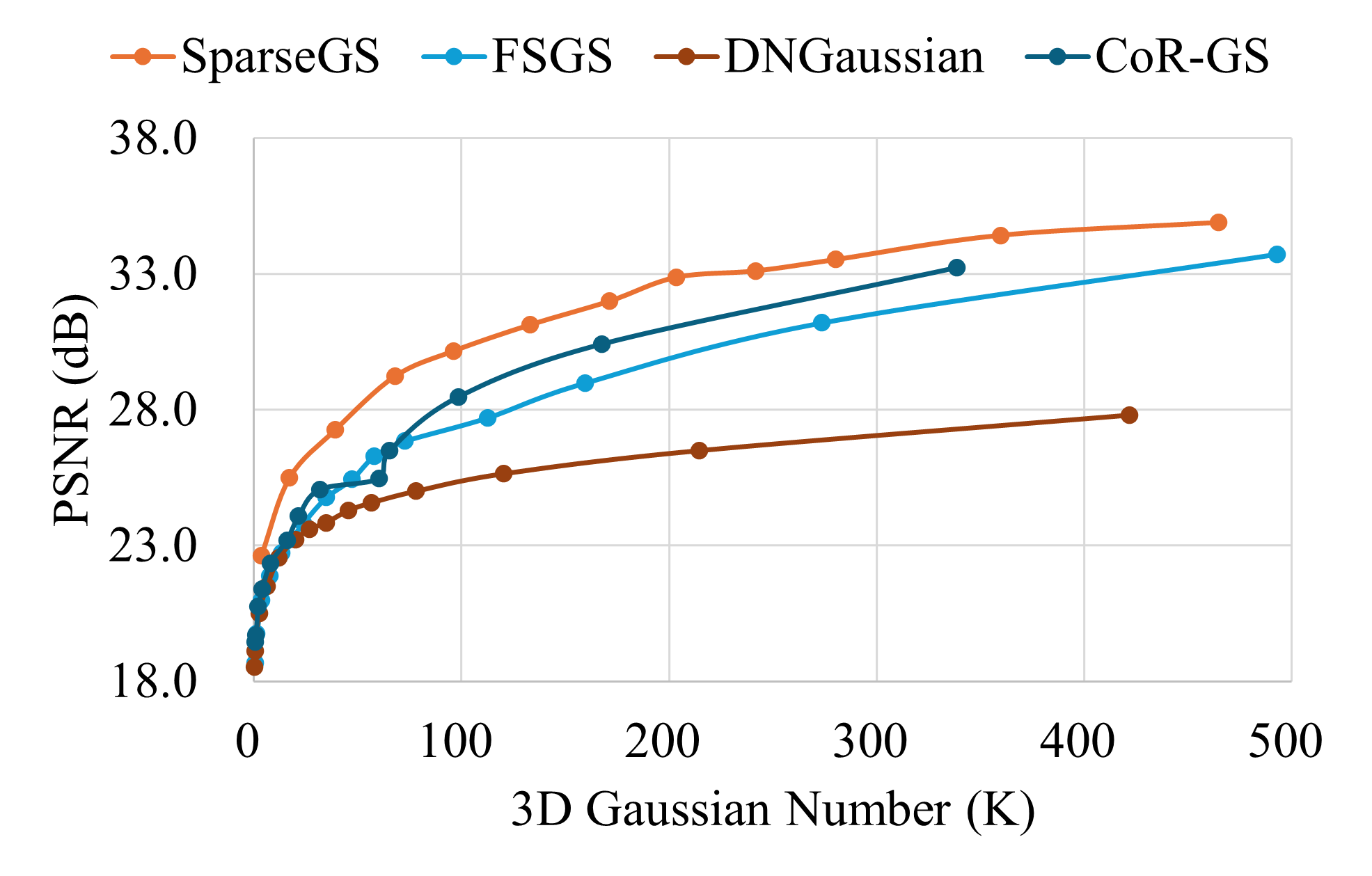}
  % \vspace{-0.4cm} 
  }
  % \hspace{-1.1em} 
    \subfloat[Test views]{
  \includegraphics[width=0.33\textwidth]{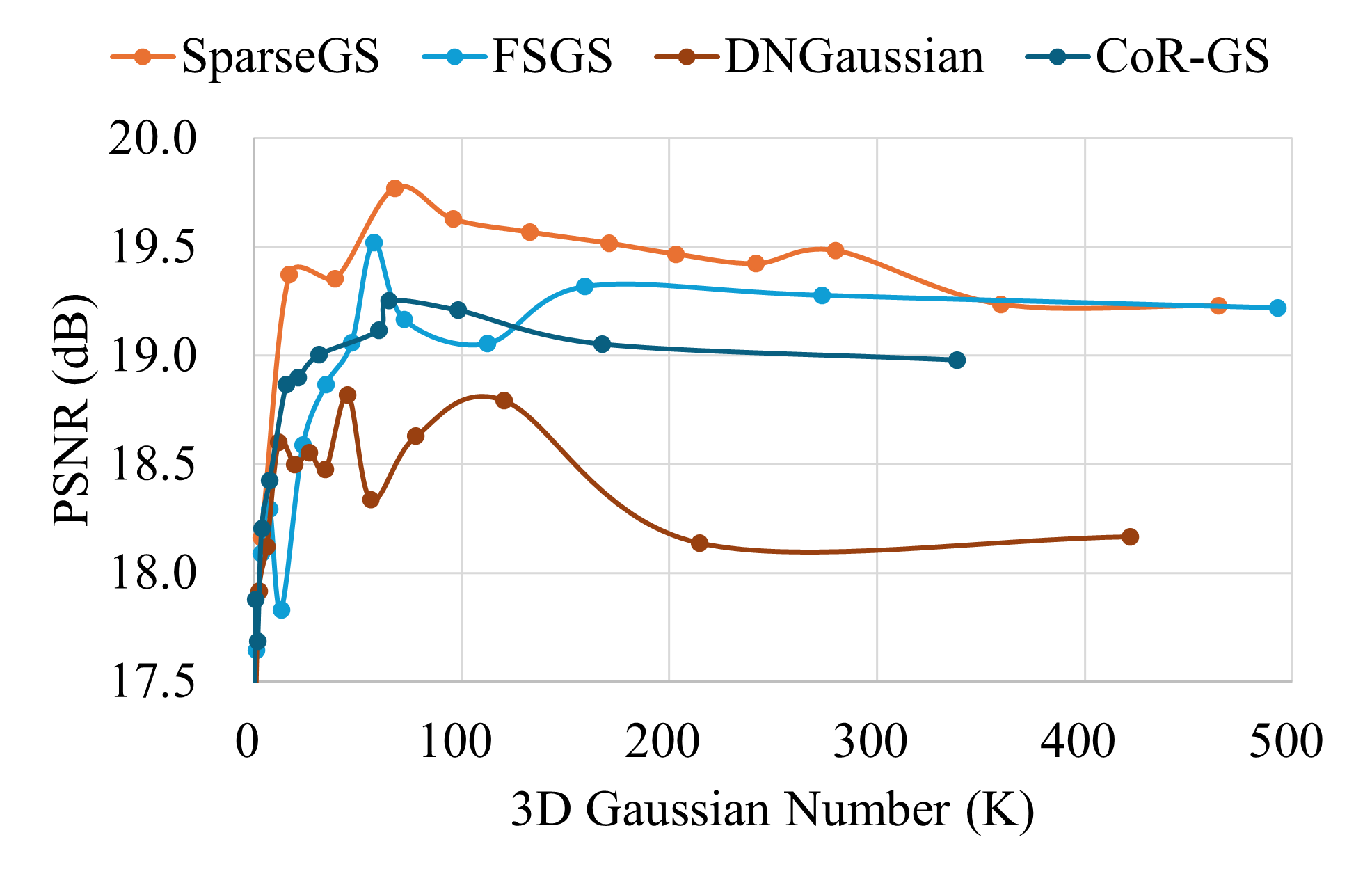}
  % \vspace{-0.4cm} 
  }
  % \hspace{-1.1em} 
    \subfloat[Validation views]{
  \includegraphics[width=0.33\textwidth]{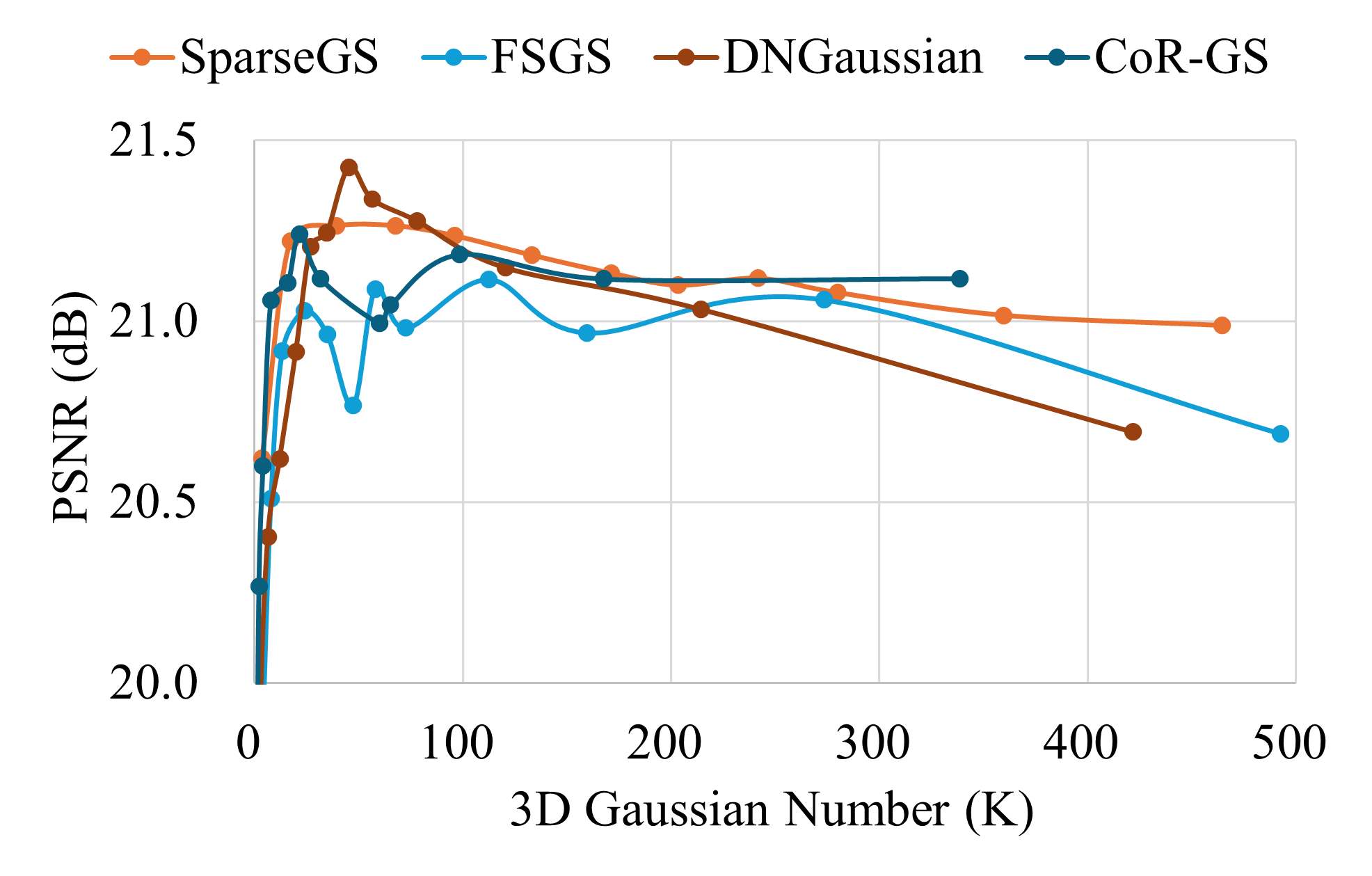}
  % \vspace{-0.4cm} 
  }
  % \hspace{-1.5em} 
  %   \subfloat[Result]{
  % \includegraphics[width=0.25\textwidth]{image/first/fir_result.png}
  % }
  \vspace{-0.4cm}
  \caption{We observe that recent sparse-view 3DGS methods \cite{li2024dngaussian, zhu2024fsgs, zhang2024cor, xiong2024sparsegs} all exhibit noticeable overfitting as the number of Gaussians increases. To address this issue, we introduce generated images as validation data into the training process for the first time. We observe that the performance trend on the validation views closely aligns with that of the test views. 
  Therefore, we propose a Validation-guided Gaussian Number Control (VGNC) strategy during training to effectively reduce overfitting.}
  \label{fig:teaser}
\end{teaserfigure}
% \received{20 February 2007}
% \received[revised]{12 March 2009}
% \received[accepted]{5 June 2009}

%%
%% This command processes the author and affiliation and title
%% information and builds the first part of the formatted document.
\maketitle

\section{Introduction}

Visual 3D reconstruction is a technique that reconstructs a scene in digital space using multi-view images. It serves as a fundamental task in various fields, including the metaverse \cite{chen2024ggavatar}, digital twins, autonomous driving \cite{li2024gs3lam}, and robotics \cite{chang2021kimera}. In recent years, the rapid development of Neural Radiance Fields (NeRF)~\cite{mildenhall2021nerf} has significantly improved 3D reconstruction, enabling highly realistic novel view synthesis. Recently, a representation known as 3D Gaussian Splatting (3DGS)~\cite{kerbl20233d} has become popular due to its significant reduction in both training time and novel view rendering time. While these methods offer exceptional performance, they need numerous densely captured images for effective scene reconstruction, which limits their practicality and widespread use.

To address this issue, recent methods~\cite{chan2024point,zhu2024fsgs,xiong2024sparsegs,li2024dngaussian,zhang2024cor} have explored leveraging 3DGS for 3D reconstruction using only sparse-view image inputs. Although these approaches have achieved certain improvements under sparse-view conditions, our experiments reveal that they still suffer from overfitting, as illustrated in Figure \ref{fig:teaser}. Limiting the maximum number of Gaussians during reconstruction results in improved rendering quality of training-view images as the number of Gaussians increases. However, for test-view images, the rendering quality initially increases, but then it declines, indicating a clear overfitting phenomenon~\cite{ying2019overview}.

To reduce the overfitting issue in sparse-view 3D Gaussian Splatting, we propose Validation-guided Gaussian Number Control (VGNC). Our core idea is to utilize the guidance from a validation set to help 3DGS automatically determine the optimal number of Gaussian points during training. This optimal point reduces overfitting, decreases Gaussian redundancy, and enhances the quality of novel view rendering. To the best of our knowledge, this is the first study to incorporate a validation set into the training process of 3DGS.

Specifically, we generate novel-view images from sparse input views using a generative NVS model. However, these synthesized views often exhibit significant distortions. Such distortions contradict the objective of 3D reconstruction, which aims to faithfully recreate real-world scenes or objects in digital space. Therefore, these distorted images should be excluded from the validation images.
To resolve this, we propose a filtering strategy to eliminate distorted images with low geometric consistency. We then perform initialization using both the filtered validation images and the original input images, resulting in an improved initial Gaussian point cloud with a better global geometric structure.
To control the number of Gaussians, we introduce a growth-and-dropout mechanism that dynamically adjusts the number of Gaussians throughout training, which helps reduce overfitting in sparse-view scenarios.

Our method can be easily integrated into various sparse-view 3DGS frameworks. We evaluate it on three widely-used and diverse datasets~\cite{barron2022mip,knapitsch2017tanks,mildenhall2019local} across four recent sparse-view 3DGS~\cite{li2024dngaussian,zhu2024fsgs,zhang2024cor,xiong2024sparsegs} and 3DGS~\cite{kerbl20233d}. For each method, we compare the standalone performance of each method with its performance when combined with our approach. Extensive experiments show that our method alleviates overfitting in sparse-view imaging, reduces memory consumption, accelerates training and rendering, and improves the quality of novel view synthesis.
In summary, the main contributions are as follows:

(1) We identify the overfitting issue in sparse-view 3DGS and, for the first time, explicitly illustrate this phenomenon through a detailed analysis.

(2) We introduce VGNC (Validation-guided Gaussian Number Control), a method that automatically identifies the optimal number of Gaussians during sparse-view 3DGS training, effectively reducing overfitting.

(3) We propose a new strategy for generating and filtering validation images using a generative NVS model. To our knowledge, this is the first attempt to introduce a validation dataset into the 3DGS.

(4) We integrate our method into five different 3DGS frameworks and evaluate it on three diverse datasets. Our method consistently identifies the optimal number of Gaussians, reduces overfitting, lowers memory usage, accelerates rendering, and improves the quality of novel view synthesis.

\begin{figure}[t]  
	\centering
	\includegraphics[width=8.5cm]{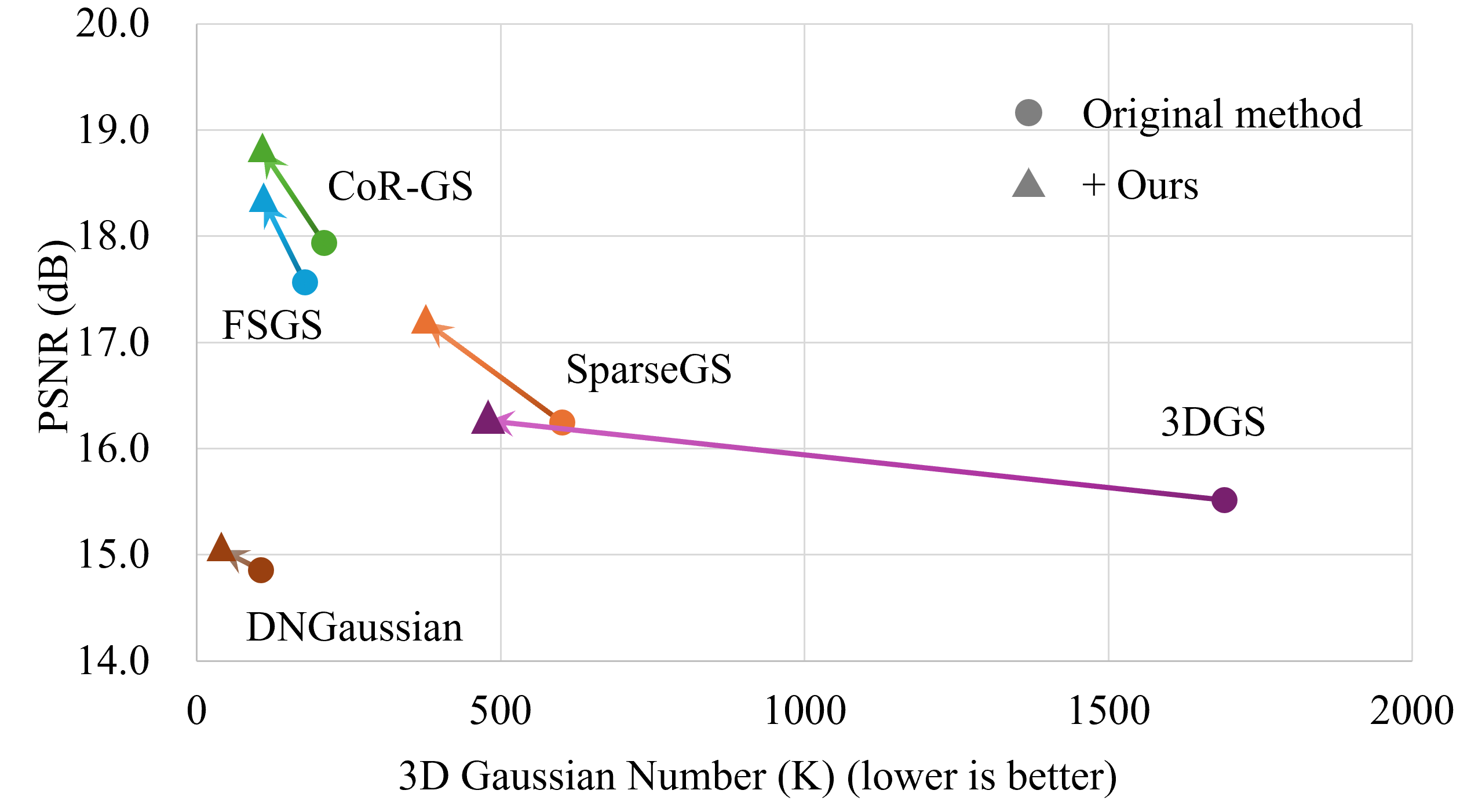}
    \vspace{-0.6cm}
	\caption{We evaluate the performance of five different methods\cite{li2024dngaussian,zhu2024fsgs,zhang2024cor,xiong2024sparsegs,kerbl20233d} on the Mip-NeRF360 dataset\cite{barron2022mip} under sparse-view reconstruction, with and without our method. Integrating our approach improves rendering quality while reducing model storage across all methods.
 }
	\label{Fig:spare_view_result}	
\end{figure}

\begin{figure*}[t]
	\centering
	\includegraphics[scale=0.43]{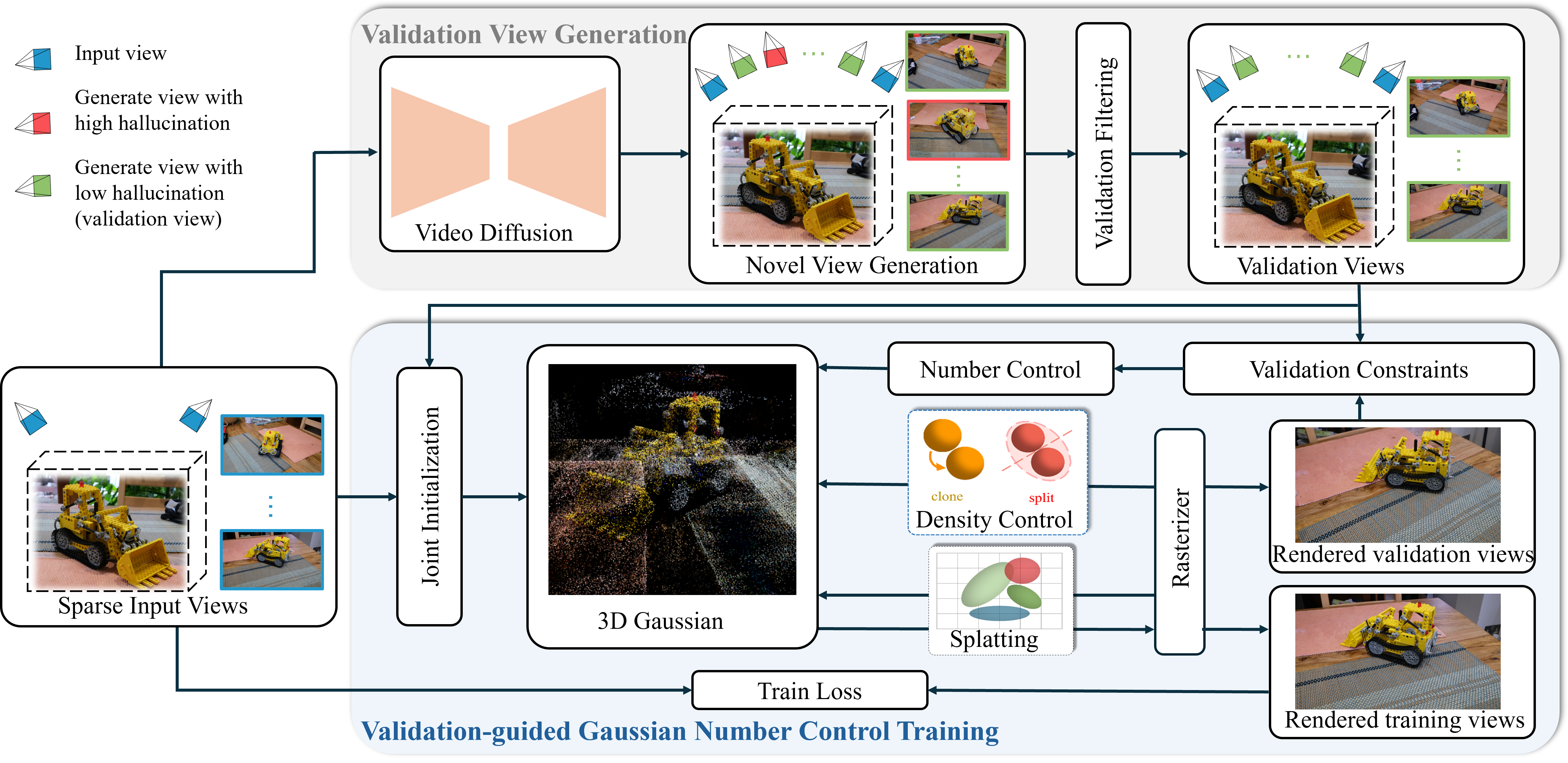}
	\caption{\textbf{Overview of VGNC.} We first propose a validation view generation method. Then, we introduce a validation-based monitor into the 3DGS training process to guide the control of the Gaussian quantity. This enables the model to automatically identify the optimal number of Gaussians during training.
	}
	\label{fig:pipeline}
\end{figure*}

\section{Related work}
\subsection{Sparse View 3D Reconstruction}
    In recent years, NeRF \cite{mildenhall2021nerf}  and its subsequent developments \cite{barron2022mip,muller2022instant,bao2023and,xuan2024superpixel} have revolutionized high-fidelity novel view synthesis in 3D reconstruction. Notably, 3DGS \cite{kerbl20233d} has significantly accelerated training and improved rendering speeds for novel viewpoints. While these methods achieve photorealistic image rendering, they typically require a dense set of input images during training. The need for hundreds of images for scene reconstruction poses a significant challenge for practical deployment. To address this limitation, various approaches have been proposed. Depth-NeRF \cite{deng2022depth} integrates depth constraints into NeRF to enhance novel view rendering quality. RegNeRF \cite{niemeyer2022regnerf} incorporates depth smoothness as a geometric regularization. FreeNeRF \cite{yang2023freenerf} introduces a dynamic frequency control module, while SparseNeRF \cite{wang2023sparsenerf} applies a spatial continuity constraint to improve performance. DNGaussian \cite{li2024dngaussian} extends 3DGS \cite{kerbl20233d} by incorporating hard and soft depth regularization to refine scene geometry. SparseGS \cite{xiong2024sparsegs} enhances 3DGS with depth and diffusion regularization to improve novel view synthesis. FSGS \cite{zhu2024fsgs} introduces a Proximity-Guided Gaussian Unpooling mechanism, significantly increasing the initial point cloud density for higher reconstruction accuracy. CoR-GS \cite{zhang2024cor} reconstructs two Gaussian representations of the same scene and minimizes their discrepancies to enhance sparse-view reconstruction. CoherentGS \cite{paliwal2024coherentgs} leverages monocular depth and flow correspondences to initialize a set of per-pixel 3D Gaussians. However, we observe that these sparse-view 3D reconstruction methods still suffer from severe overfitting, as illustrated in Figure \ref{fig:teaser}. To address this issue, we propose ValuationGaussian, a novel approach designed to mitigate overfitting in sparse-view 3D reconstruction.

  \subsection{Image Generation via Diffusion Models}
In recent years, diffusion models \cite{ho2020denoising,songdenoising,yang2024hi3d} have demonstrated remarkable capabilities in generating high-quality images. These methods \cite{rombach2022high} can generate novel-view images based on textual prompts or reference images, highlighting their potential for extension into 3D space. DreamFusion \cite{pooledreamfusion} employs a pre-trained diffusion model to guide 3D object generation from prompts. Zero-1-to-3 \cite{liu2023zero} introduces a camera pose-conditioned diffusion model trained on synthetic datasets \cite{deitke2023objaverse,chang2015shapenet}, enabling the generation of novel-view images of single objects against simple backgrounds. ZeroNVS \cite{sargent2024zeronvs} enhances Zero-1-to-3 \cite{liu2023zero} by improving its ability to generate objects in complex backgrounds using a dataset that combines synthetic \cite{deitke2023objaverse} and real-world images \cite{zhou2018stereo,reizenstein2021common,yu2023mvimgnet}, allowing novel views to be synthesized from a single reference image. ReconFusion \cite{wu2024reconfusion} leverages images generated by a 2D diffusion model to train PixelNeRF \cite{yu2021pixelnerf}. ViewCrafter \cite{yu2024viewcrafter} integrates a video diffusion model \cite{xing2024dynamicrafter} with point cloud reconstruction to generate new views of a scene. 3DGS-Enhancer \cite{liu20243dgs} employs a video diffusion model \cite{blattmann2023stable} and trajectory interpolation to enhance unbounded 3DGS representations.
Although these methods enrich 3D content, they fail to address a fundamental issue: the inconsistency between the real world and generated images, which often contain hallucinated details. Since 3D reconstruction aims to create an accurate digital replica of reality, it is crucial to avoid the inclusion of such hallucinated content. Unlike previous approaches, our method uses diffusion-generated images only for validation and initialization, rather than constraining the 3D reconstruction process, thereby preventing hallucinated artifacts from interfering with the reconstruction.

\subsection{Overfitting}

  Overfitting is a common issue in supervised machine learning \cite{ying2019overview}, typically characterized by a model performing increasingly well on training data while exhibiting deteriorating performance on unseen test data. As illustrated in Figure \ref{fig:teaser}, we observe a typical overfitting phenomenon in sparse-view 3D reconstruction. Overfitting can generally be mitigated through dataset expansion, additional regularization constraints, model reduction strategies, or early stopping techniques. ViewCrafter \cite{yu2024viewcrafter} and 3DGS-Enhancer \cite{liu20243dgs} expand novel view datasets using video diffusion models. However, current diffusion models often generate hallucinated content that deviates from real-world structures, contradicting the goal of digitally reconstructing real environments. At the same time, some studies \cite{paliwal2024coherentgs,zhang2024cor, zhu2024fsgs,li2024dngaussian,xiong2024sparsegs} introduce various new regularization constraints, and severe overfitting issues persist, as shown in Figure \ref{fig:teaser}. Unlike previous approaches, our method simultaneously addresses overfitting in sparse-view 3D reconstruction by introducing a validation set, employing model reduction strategies, and implementing early stopping techniques. Our approach not only achieves superior novel view rendering quality but also significantly reduces model storage requirements.

\section{Method}

Figure \ref{fig:pipeline} presents an overview of the proposed ValidationGaussian, which can be easily integrated into most 3DGS frameworks \citep{kerbl20233d}. The method introduces validation set supervision to prevent overfitting in 3DGS training, particularly under sparse-view conditions.
In this section, we first provide a brief background on 3DGS. Then, we present a detailed discussion of the motivation for introducing a validation set. Finally, we detail the implementation of our approach, including validation set generation and filtering, joint initialization, validation-guided monitor, number control, and a novel adaptive gaussian training strategy.

\subsection{Preliminary: 3D Gaussian Splatting}
\label{sec:3DGS}
3D Gaussian Splatting (3DGS) \citep{kerbl20233d} is a recently developed method for 3D scene reconstruction that models a scene using numerous anisotropic 3D Gaussians. This formulation is fully differentiable, enabling the use of learning-based optimization strategies, while also incorporating explicit spatial encoding, which facilitates intuitive scene manipulation and editing. Moreover, it supports rapid, rasterization-style rendering via splatting, achieving both speed and efficiency.

The reconstruction pipeline begins with an unordered collection of scene images, which are processed through Structure-from-Motion (SfM) \cite{schoenberger2016sfm} to estimate camera poses and generate a sparse point cloud. These sparse points are then utilized by 3DGS to initialize the mean $\mu$ of each 3D Gaussian:
\begin{equation}
    G(x)=e^{-\frac{1}{2}(x-\mu)^{T} \Sigma^{-1}(x-\mu)}
\end{equation}
where $\Sigma$ represents the covariance matrix of the 3D Gaussian, and $x$ denotes any point in the 3D scene.
% $\Sigma$ is defined using a scaling matrix $\boldsymbol{S}$ and a rotation matrix 
% $\boldsymbol{R}$:

% \begin{equation}
%     \Sigma=\boldsymbol{R} \boldsymbol{S} \boldsymbol{S}^{T} \boldsymbol{R}^{T}
% \end{equation}

The 3D Gaussian function $G(x)$ is projected onto the image plane via the camera’s intrinsic matrix, resulting in a corresponding 2D Gaussian distribution. The final image is then synthesized using alpha blending:

\begin{equation}
    C\left(x^{\prime}\right)=\sum_{k \in Q} c_{k} \alpha_{k} \prod_{j=1}^{k-1}\left(1-\alpha_{j}\right)
\end{equation}
where $x^{\prime}$ denotes the target pixel location, $Q$ is the total number of 2D Gaussians influencing that pixel, and $\alpha$ represents the opacity of each individual Gaussian. The color $c$ associated with each Gaussian is expressed using spherical harmonics.
All attributes of the 3D Gaussians are learnable and optimized directly in an end-to-end manner during training.

\begin{figure}[t]  
	\centering
	%	\quad
	\subfloat[Coefficient and number]{
		\includegraphics[width=2.8cm]{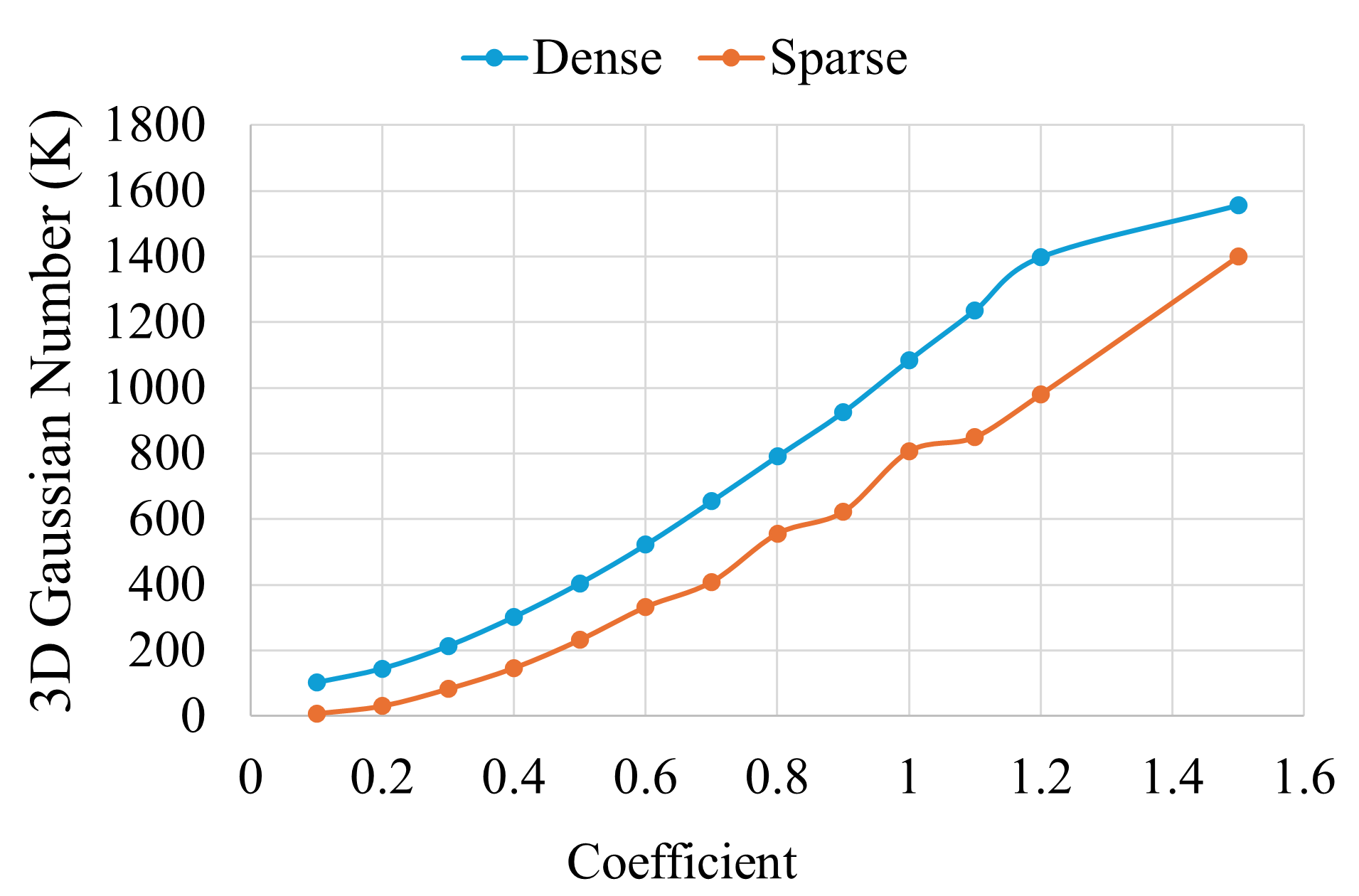}
	}
      % \hspace{-1.0em} 
	\subfloat[Train view]{
		\includegraphics[width=2.8cm]{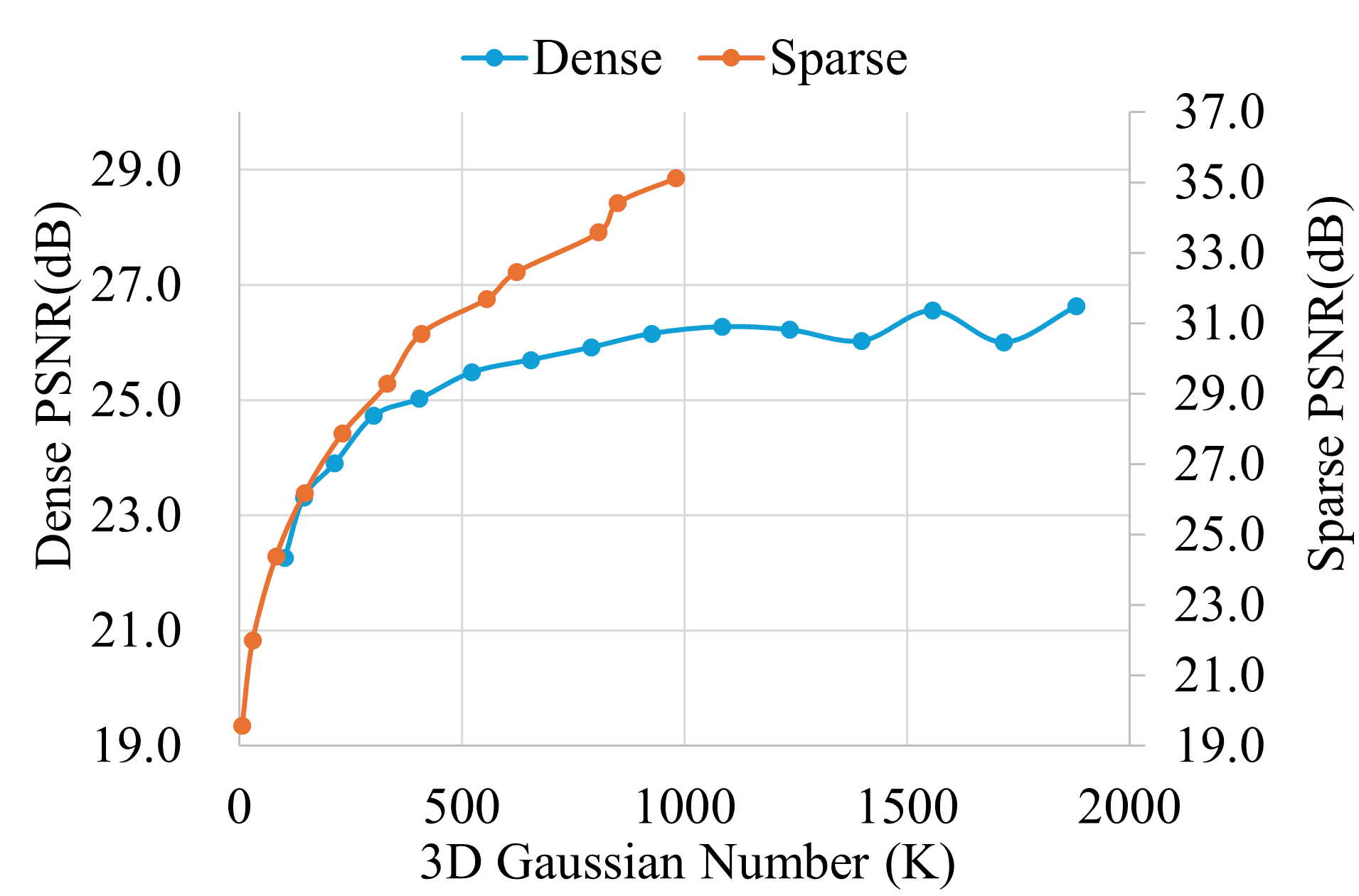}
	}
      % \hspace{-1.0em} 
        \subfloat[Test view]{
		\includegraphics[width=2.8cm]{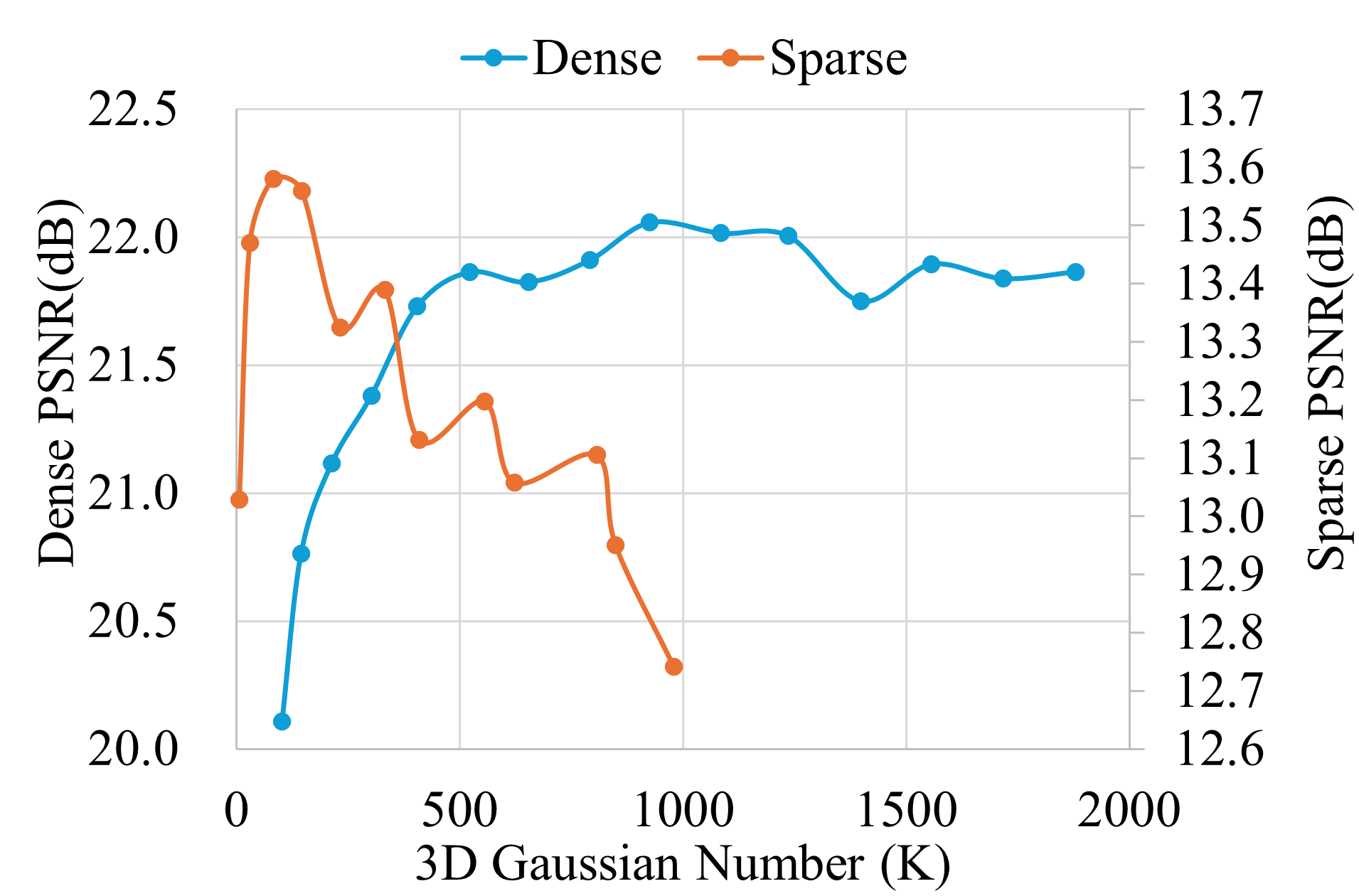}
	}
      \vspace{-0.4cm}
	\caption{ Gaussian number vs. other variables.
 }
	\label{Fig:num_curve}	
\end{figure}

\subsection{Motivation of Validation}

Inspired by the multimodal regularization in ThermalGaussian \cite{lu2024thermalgaussian}, we observed that the coefficient applied before the loss function in 3DGS is proportional to the number of Gaussians, as shown in Figure \ref{Fig:num_curve} Additionally, we identified an interesting phenomenon: as the number of Gaussians in 3DGS increases, the rendering quality of images from training viewpoints continues to improve. However, the rendering quality of images from test viewpoints eventually saturates. This is a classic overfitting phenomenon \cite{prechelt2002early}. This effect is particularly pronounced in sparse-view 3DGS, where test-view rendering quality initially increases but then declines, as illustrated in Figure \ref{Fig:num_curve}.

Addressing overfitting is a fundamental challenge in supervised machine learning \cite{ying2019overview}, while simultaneously reducing computational overhead and improving reconstruction quality remains a persistent goal. As shown in Figure \ref{Fig:num_curve}, a distinct non-overfitting point exists in sparse-view 3DGS, where both rendering quality is higher and memory usage is lower. To enable sparse-view 3DGS and related algorithms to automatically identify this optimal Gaussian numbers during reconstruction, we introduce a validation-guided Gaussian number control approach.

\begin{figure}[t]  
	\centering
	%	\quad
	\subfloat[ViewCrafter's input view]{
		\includegraphics[width=4.0cm]{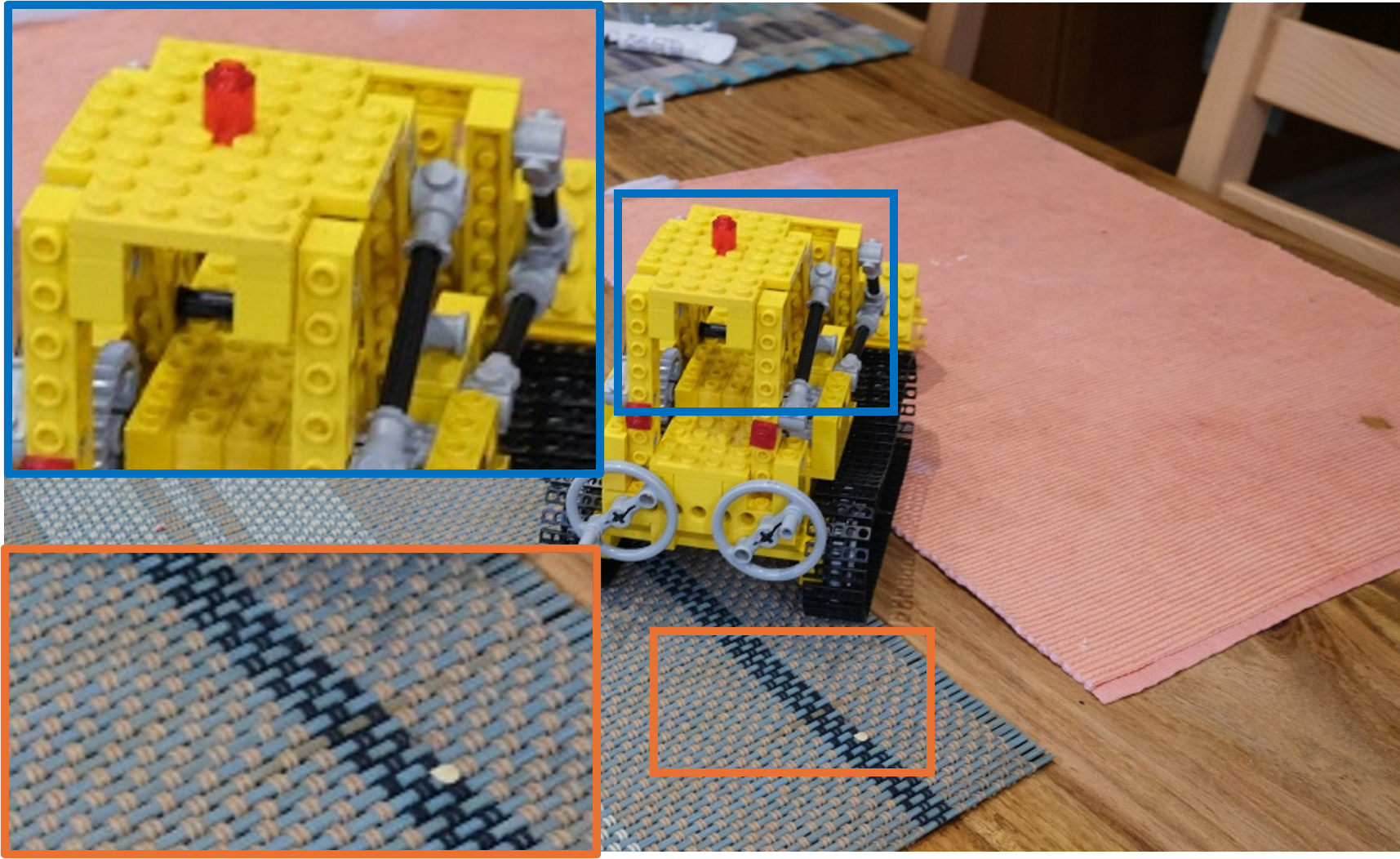}
	}
	\subfloat[Low-hallucination generation view]{
		\includegraphics[width=4.0cm]{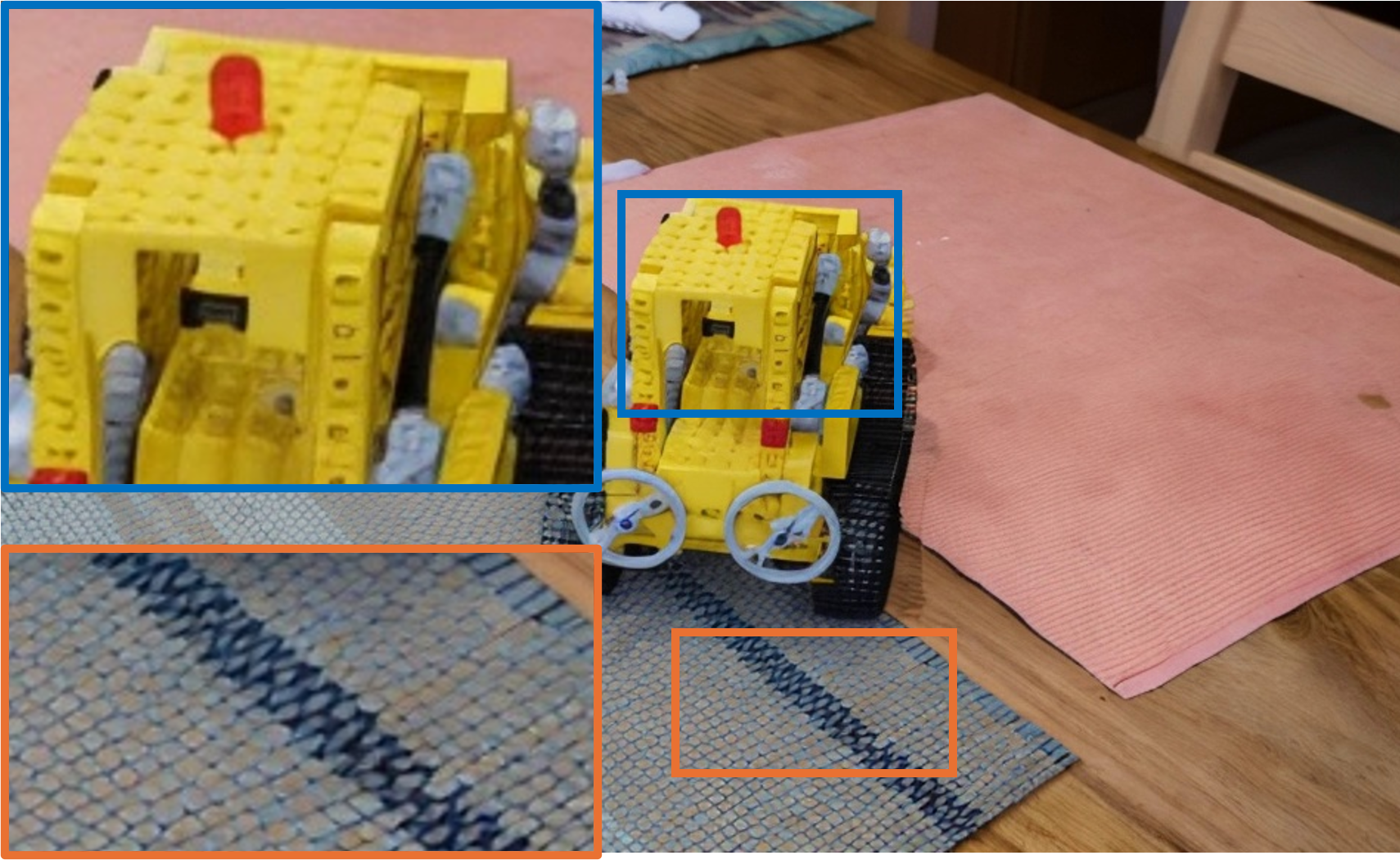}
	}
    \\
        \subfloat[Real reference view for (d)]{
		\includegraphics[width=4.0cm]{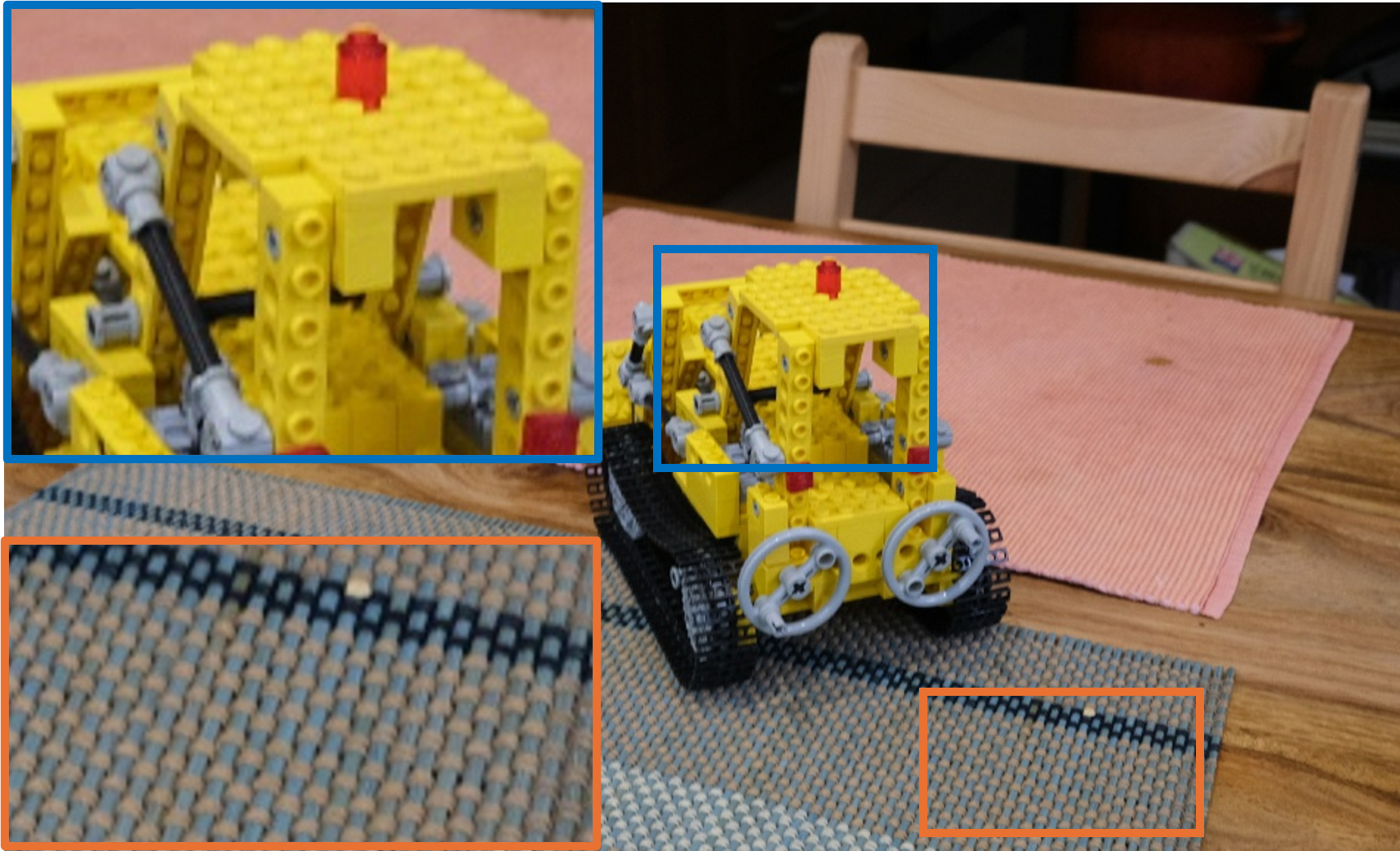}
	}
        \subfloat[High-hallucination generation view]{
		\includegraphics[width=4.0cm]{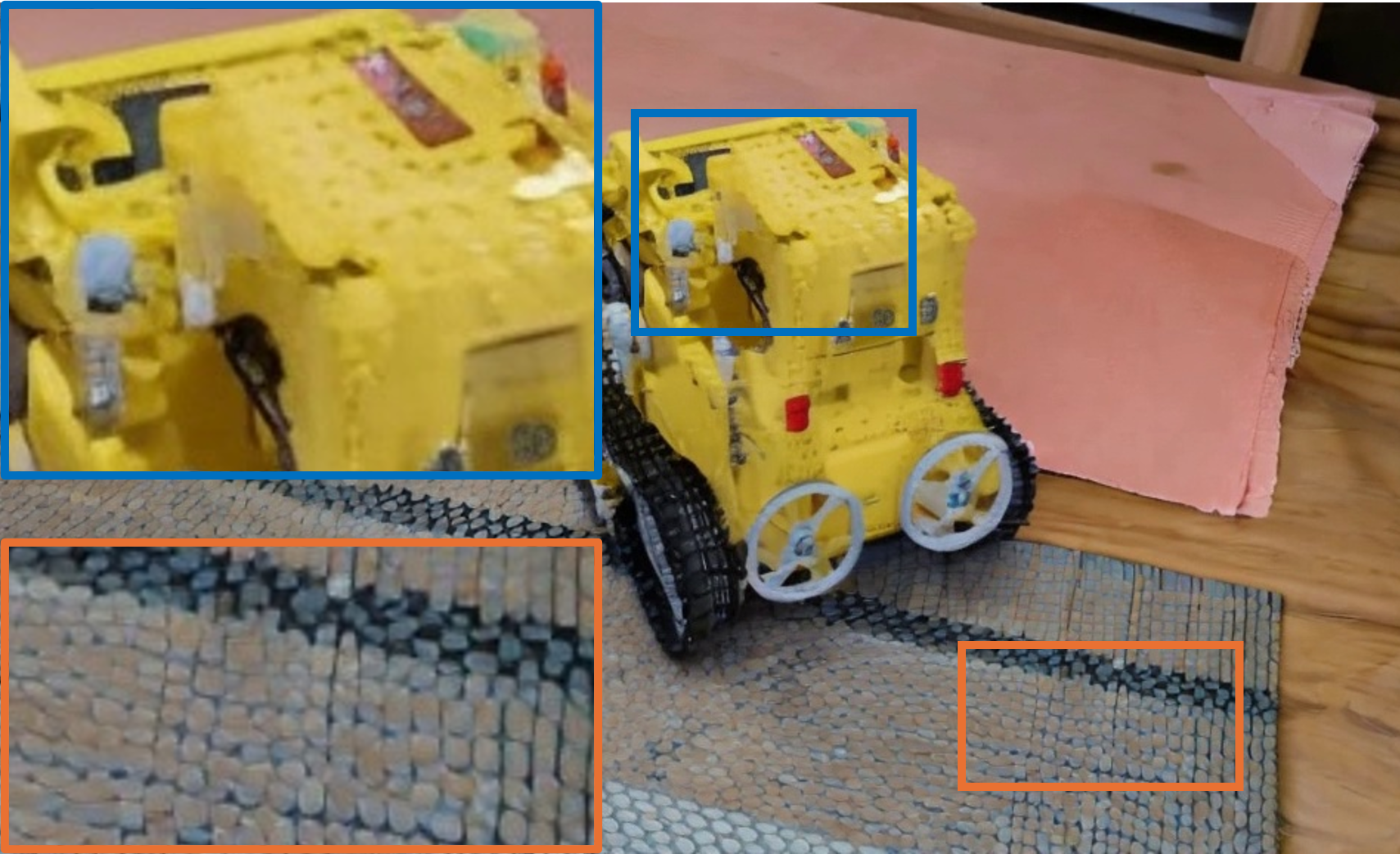}
	}
    \vspace{-0.4cm}
	\caption{Generated views with differing hallucinations.
 }
	\label{Fig:hallu}	
\end{figure}

\subsection{Validation Images Generation}
 \label{sec:validation generation}

To generate effective validation images for 3D reconstruction from the $n$ input sparse-view images $\{I_{1}, I_{2}, \cdots, I_{n}\}$, we first employ a diffusion model to generate novel-view images:

\begin{equation} 
    \{J_{1}, J_{2}, \cdots, J_{m}\}
     = \mathcal{F}(\{I_{1}, I_{2}, \cdots, I_{n}\})
\end{equation}
where $\mathcal{F}(\cdot)$ represents ViewCrafter \cite{yu2024viewcrafter}, a novel video diffusion model designed for multi-view consistency. Here, $m$ represents the number of generated images. In ViewCrafter\cite{yu2024viewcrafter}, we generate 25 new viewpoints between each pair of input images, resulting in $m=25(n-1)$.
Although recent video diffusion models can generate high-quality novel views with a certain degree of multi-view consistency, these images still contain hallucinations that differ from real-world details. As shown in Figure \ref{Fig:hallu}(b), even novel views that are very close to the input viewpoints contain minor hallucinations. In Figure \ref{Fig:hallu}(d), some views exhibit severe hallucinations.

The goal of 3D reconstruction is to accurately reproduce real-world objects or scenes in digital 3D space. Therefore, generated images with severe hallucinations cannot be directly used as training constraints, as the hallucinations compromise the realism of the reconstruction. These severely hallucinated images also cannot serve as validation inputs, since their unrealistic content negatively affects quantitative evaluation. To prevent hallucinations from interfering with validation-guided monitor, we design a validation image filtering method to remove generative images with significant hallucinations.

We first extract Scale-invariant Feature Transform (SIFT) \cite{lowe1999object} points and descriptors from all input images$\{I_{1}, I_{2}, \cdots, I_{n}\}$ and all generated images$\{J_{1}, J_{2}, \cdots, J_{m}\}$. Next, we compute confidence maps between each generated image and all input images. Based on these results, we select the generated images with minor hallucinations to form the validation set. For any given pair of a generated image $G_{j}$ and an input image $I_{i}$, we apply the FLANN \cite{muja2009fast} algorithm for fast feature matching to obtain a set of corresponding feature point pairs. These pairs are then processed using the RANSAC \cite{camposeco2018hybrid} algorithm to estimate the essential matrix $\mathbf{E}$, which encodes the relative rotation and translation between the input and generated views:

\begin{equation} 
    \mathbf{E}=[\mathbf{t}]_{\times} \mathbf{R}
\end{equation}
where $[\mathbf{t}]_{\times}$  is the skew-symmetric matrix of the translation vector $\mathbf{t}$. The vector $\mathbf{t} \in \mathbb{R}^{3 \times 1}$ represents the translation between the two viewpoints, while the matrix $\mathbf{R} \in \mathbb{R}^{3 \times 3}$ denotes the relative rotation between them.
By performing Singular Value Decomposition (SVD) \cite{hartley2003multiple} on the essential matrix $\mathbf{E}$, we solve for $\mathbf{R}$ and $\mathbf{t}$. Using the camera intrinsic matrix $\mathbf{K}$ and the rotation and translation between the two cameras, we can compute the projected pixel position $(\tilde{u}, \tilde{u})$ in the input view for each pixel position $(u, v)$ in the generated view:

\begin{equation}
\left[\begin{array}{c}
\tilde{u} \\
\tilde{v} \\
1
\end{array}\right] = \frac{1}{\tilde{z}} \mathbf{K}\left(\mathbf{R}\cdot \mathbf{K}^{-1} \left[\begin{array}{c}
u \\
v \\
1
\end{array}\right]+ \boldsymbol{t}\right)
\end{equation}
where $\tilde{z}$ is the normalization factor.
Next, using the pixel-wise transformation defined in Equation (5), we back-project the pixel content from the generated image $J_j$ onto the input view $I_i$, resulting in a reprojected image $\tilde{I}_{i \leftarrow j}$:

\begin{equation}
\tilde{I}_{i \leftarrow j}(\tilde{u}, \tilde{v}) = J_{j}(u, v)
\end{equation}

Since some pixels in the generated image $J_j$ may be projected outside the bounds of the input view $I_i$, or multiple pixels may be projected to the same location, the reprojected image $\tilde{I}_{i \leftarrow j}$ may contain some empty pixels without valid values. We apply bilinear interpolation to fill in these small blank regions.

We then compute a confidence map by measuring the pixel-wise differences between the reprojected image and the original input image:
\begin{equation}
M_{i \leftarrow j}\left(u, v\right)=\exp \left(-\frac{\left\|I_{i}\left(u, v\right)-\tilde{I}_{i \leftarrow j} (u, v)\right\|^{2}}{\sigma^{2}}\right)
\end{equation}
Where $\left\|\cdot\right\|$represents the Euclidean distance of pixel values, and $\sigma$ is the scale parameter for confidence attenuation.
When the hallucination content is minimal or absent, the re-projection error remains small, resulting in a high confidence score. Conversely, the hallucination content is severe, the re-projection error becomes large, leading to a lower confidence score.
Therefore, we determine whether a generated image contains significant hallucinations by counting the number of low-confidence pixels in its reprojected views. The number of low-confidence pixels $N_{i \leftarrow j}$ for a generated image $G_j$ on an input view $I_i$ is defined as:
\begin{equation}
N_{i \leftarrow j} = \sum_{u=1}^{h} \sum_{v=1}^{w} \mathbb{I}(M_{i \leftarrow j}(u,v) \leq  \theta)
\end{equation}
where $\mathbb{I}(\cdot)$ is the indicator function, which returns 1 if the condition inside the brackets is true, and 0 otherwise:
\begin{equation}
\mathbb{I}(M_{i \leftarrow j}(u,v) \leq  \theta) = 
\begin{cases} 
1 & \text{if } M_{i \leftarrow j}(u,v) \leq  \theta, \\
0 & \text{otherwise}.
\end{cases}
\end{equation}

We observe that if the viewpoint difference between the generated image and an input image is too large—such that they may not even depict the same object—the confidence score drops significantly.
To address this, we compute the number of low-confidence pixels between each generated image $J_j$ and all input images $\{I_1, I_2, \dots, I_n\}$, resulting in a set $\{N_{1 \leftarrow j}, N_{2 \leftarrow j}, \dots, N_{n \leftarrow j}\}$.
Among these, we select the view corresponding to the minimum value $N_j^{\min}$ as the closest input view to the generated image $J_j$.
If this closest input view still contains a large number of low-confidence pixels when compared with $J_j$, we consider $J_j$ to contain excessive hallucinations and discard it.
The final validation images $\{V_{1}, V_{2},\dots, V_{p}\}$ are constructed by retaining only the generated images that pass this filtering criterion, defined as:
\begin{equation}
\left\{V_{1}, V_{2}, \cdots, V_{p}\right\} = \left\{J_{j} \mid  N^{min}_{j} \leq \tau\right\}
\end{equation}

\subsection{Validation-guided Gaussian Training}
\textbf{Joint  Initialization.}
Recent works have improved the performance of sparse-view 3DGS by enhancing the 3D geometric information extracted from limited input images. For example, some methods introduce depth maps to strengthen the initial point cloud \cite{paliwal2024coherentgs}, add depth constraints \cite{wang2023digging} during optimization \cite{xiong2024sparsegs}, or densify the initial Gaussians \cite{zhu2024fsgs,zhang2024cor}.
In this study, we introduce validation images, which are novel views with initial geometric consistency and effectively extend the input images. Although these generated validation images still contain minor hallucinations in certain details, they provide a significant amount of correct geometric structure and contain richer potential 3D geometric information. For instance, depth estimated from multi-view geometry methods is often more accurate than that from monocular methods such as Depth Anything \cite{yang2024depth}.
Therefore, we perform joint initialization using both the sparse input images and the validation images generated in Section \ref{sec:validation generation}. This process produces an initial point cloud with richer geometric information, leading to improved final reconstruction performance.

\begin{algorithm}[b]
\caption{Validation-guided Gaussian Number Control (VGNC)}
\begin{algorithmic}  % [1] 显示行号
% \Require int $a, b$($a > 0, b > 0$)
\State $V=(V_{1}, V_{2}, \cdots, V_{p}) \leftarrow $ Validation Images Generation (Sec. \ref{sec:validation generation})
\State $i \leftarrow 0 $ \Comment{Iteration Count}
\State $\mathcal{M}^{opt}\leftarrow 1$ \Comment{Optimal Validation-guided Monitor}
\While{not converged}
    \State $\mathcal{L}_{original} = \mathcal{L}(I,\hat{I})$ \Comment{Original train Loss}
    \State Optimization()
    \If{$i < DensificationIteration(i)$}
        \State GaussianDensity() \Comment{Number Increase}
           \If{$i \bmod ValidationIteration(i) = 0$}
        \State $\mathcal{M}=\frac{1}{p} \sum_{k=1}^{p}\left\|V_k-\hat{V}_{k}\right\|_{2}^{2}$ \Comment{According to Eq.\ref{eq:val_monitor}}
        \If{$\mathcal{M} < \mathcal{M}^{opt}$}
        \State $\mathcal{M}^{opt} = \mathcal{M}$
        \State $Num^{opt} = $ GaussiansShape() \Comment{Optimal Number}
        
        \EndIf
        
    \EndIf
    \Else 
    % \State $\beta = Num^{opt}$ 
    \State GaussianDropout() \Comment{ Number Reduction}
    \If{GaussiansShape() $< Num^{opt}$}
        \State GaussianDensity()
    \EndIf     
        
    \EndIf
    \State $i \leftarrow i + 1$
\EndWhile
\end{algorithmic}
\end{algorithm}

\noindent \textbf{Validation-guided Monitor.}
Although these validation images contain minor hallucinations, using them as training inputs could severely distort the reconstruction results. However, when used solely for guiding quantitative evaluation, the noise is negligible at the image level and thus acceptable. Therefore, we treat the generated validation images as ground truth and compute the validation-guided monitor between them and the rendered images $\hat{V}_{k}$ from 3DGS at the validation viewpoints:
\begin{equation}
% \mathbb{C}
\mathcal{M}=\frac{1}{p} \sum_{k=1}^{p}\left\| V_k-\hat{V}_{k}\right\|_{2}^{2}
\label{eq:val_monitor}
\end{equation}
where $p$ represents the number of images in the validation set. 
Since the PSNR and the validation-guided monitor $\mathcal{M}$ exhibit a consistent inverse correlation, we monitor the change in $\mathcal{M}$ during training to determine whether overfitting occurs and to identify the optimal number of Gaussians.

\noindent \textbf{Number Control.}
We achieve precise control over the total number of Gaussian points by adjusting the loss coefficient, Gaussian dropout, and total count threshold. As shown in Figure \ref{Fig:num_curve}, increasing the pre-loss coefficient during training leads to the generation of more Gaussians. However, this method alone cannot precisely regulate the total number of Gaussians. To address this, we introduce a Gaussian count threshold: once the number of Gaussians reaches this threshold, cloning and splitting operations are halted, ensuring precise control over the final Gaussian count. Additionally, Gaussian dropout is employed during iterative training when the number of Gaussians significantly exceeds the optimal count. This technique rapidly removes a large number of points at random, allowing the reconstruction process to efficiently revert to the optimal non-overfitting Gaussian count.

\noindent\textbf{Training Strategy.}
As shown in Figure \ref{fig:teaser}, the optimal Gaussian number point on the test set closely aligns with that on the validation set. To prevent overfitting in 3DGS training, we propose a novel training paradigm that integrates validation guidance and Gaussian number control.

As outlined in Algorithm 1, our training strategy begins by setting an initially low Gaussian count threshold. After a certain period of training, we record the validation constraint $V_{1}$. We then progressively increase the point cloud threshold while continuously recording the corresponding validation constraint $V_{k}$.
Based on the relationship between PSNR and L2 loss shown in Figure \ref{fig:teaser}, as the number of points increases, $V_{k}$initially decreases but eventually stabilizes or starts increasing. When rises continuously for several iterations, we identify this as the onset of overfitting in 3DGS training. At this point, we apply Gaussian dropout to significantly reduce the number of points, reverting to the count just before $V_{k}$reached its minimum. We then set the final Gaussian count threshold to the number of points corresponding to the lowest recorded $V_{k}$.
This approach enables 3D reconstruction to adaptively determine the optimal Gaussian number during training through validation guidance.

\vspace{-0.4cm}

\section{Experiments}

\subsection{Experimental Settings}
\textbf{Datasets.}
We conducted experiments on four different datasets. LLFF Datesets \cite{mildenhall2019local} contains 8 forward-facing real-world scenes. Following FSGS \cite{zhu2024fsgs}, we use 3 views for training in all methods. The remaining images are used for testing. 
% NeRF-Synthetic Datasets (Blender)\cite{mildenhall2021nerf} consist of multi-view images of 8 photorealistic objects rendered in Blender. These scenes have clean backgrounds and focus on a single object. We select 8 views for training and use the rest for testing. 
Mip-NeRF360 Datasets \cite{barron2022mip} include 9 real-world scenes, each containing a complex central object and detailed background. This dataset better reflects real-world small-scale reconstruction tasks. Following SparseGS \cite{xiong2024sparsegs}, we use 12 views for training and the rest for testing. 
Tanks $\&$ Temples Datasets \cite{knapitsch2017tanks} provide real-world data for large-scale scene reconstruction. Sparse-view reconstruction on such large scenes is especially challenging. Following 3DGS \cite{kerbl20233d}, we select two scenes for evaluation. We use 24 views for training and the remaining for testing.
To support view generation and validation image selection using ViewCrafter \cite{yu2024viewcrafter}, we resize all images in these datasets to 1024×576, which is the fixed resolution in ViewCrafter.

% \textbf{LLFF Datasets\cite{mildenhall2019local}}

% \textbf{NeRF-Synthetic Datasets (Blender)\cite{mildenhall2021nerf}}

% \textbf{Mip-NeRF360 Datasets\cite{barron2022mip}}

% \textbf{Tanks $\&$ Temples Datasets\cite{knapitsch2017tanks}}

\noindent \textbf{Comparison Baselines.}
We validate that our guided Gaussian Number Control can be seamlessly integrated into various 3DGS frameworks. In sparse-view 3DGS settings, our method effectively identifies the optimal number of Gaussians. To demonstrate this, we incorporate our method into four advanced sparse-view 3DGS approaches: FSGS \cite{zhu2024fsgs}, SparseGS \cite{xiong2024sparsegs}, DNGaussian \cite{li2024dngaussian}, and CoR-GS \cite{zhang2024cor}. We compare their performance with and without our method across the four benchmark datasets described above. Additionally, we evaluate the performance of the original 3DGS \cite{kerbl20233d} method and its combination with our approach, under both dense-view and sparse-view reconstruction settings.

\noindent \textbf{Evaluation Metrics.}  
Similar to 3DGS, We present quantitative evaluations of the reconstruction performance through reporting metrics including Peak Signal-to-Noise Ratio (PSNR) \cite{hore2010image}, Structural Similarity Metric (SSIM) \cite{wang2004image}, and Learned Perceptual Image Patch Similarity (LPIPS) \cite{zhang2018unreasonable}. In addition, we report the number of Gaussians required, the training time, and the novel view rendering time for different methods.

\begin{table*}[!t]
	\centering
	% \begin{tabular}{ccccccccc} 
 	\caption{ We conduct a quantitative comparison of four recent sparse-view 3DGS methods and their corresponding versions integrated with our VGNC across three different datasets under sparse-view settings. Here, ‘--’ denotes the original version of each method without our VGNC.
 } 
        \begin{tabular}{lc|cccc|cccc|cccc}  
	\toprule[2pt]
            \multicolumn{2}{c}{\multirow{2}{*}{Methods}}
            &\multicolumn{4}{c}{LLFF\cite{mildenhall2019local}}
		&\multicolumn{4}{c}{Tanks\cite{knapitsch2017tanks}}
            &\multicolumn{4}{c}{Mip-NeRF360\cite{barron2022mip}}\\
        \cmidrule(lr){3-6} \cmidrule(lr){7-10} \cmidrule(lr){11-14} 
            & & PSNR$\uparrow$  & SSIM$\uparrow$ & LPIPS$\downarrow$ & Num.$\downarrow$ & PSNR$\uparrow$  & SSIM$\uparrow$ & LPIPS$\downarrow$ & Num.$\downarrow$ & PSNR$\uparrow$  & SSIM$\uparrow$ & LPIPS$\downarrow$ & Num.$\downarrow$ \\
	
		\toprule[1pt]

            % \toprule[0.1pt]
            \multirow{2}{*}{FSGS\cite{zhu2024fsgs}}& -- & 16.878 & 0.475 & 0.434 & 229K & 16.586 & 0.570 & 0.455 & 188K & 17.567 & 0.468 & 0.513 & 178K\\
            &\textbf{+Ours} & \textbf{18.111} & \textbf{0.554} & \textbf{0.392} & \textbf{109K} & \textbf{17.341} & \textbf{0.630} & \textbf{0.391} & \textbf{166K} & \textbf{18.336} & \textbf{0.505} & \textbf{0.490} & \textbf{110K}
            \\
            \toprule[0.5pt]
            % \specialrule{0.3pt}{0pt}{0pt}
            \multirow{2}{*}{CoR-GS\cite{zhang2024cor}}& -- & 16.624 & 0.483 & 0.437 & 94K & 16.450 & 0.578 & 0.441 & 191K & 17.933 & 0.486 & 0.514 & 209K\\
            &\textbf{+Ours} & \textbf{18.314} & \textbf{0.576} & \textbf{0.395} & \textbf{57K} & \textbf{17.590} & \textbf{0.643} & \textbf{0.387} & \textbf{136K} & \textbf{18.802} & \textbf{0.526} & \textbf{0.477} & \textbf{109K}
            \\
            \toprule[0.5pt]
            \multirow{2}{*}{DNGaussian\cite{li2024dngaussian}}& -- & 16.439 & 0.476 & 0.473 & 31K & \textbf{13.148} & \textbf{0.498} & 0.594 & \textbf{21K} & 14.855 & 0.397 & 0.612 & 106K \\
            & \textbf{+Ours} & \textbf{17.755} & \textbf{0.532} & \textbf{0.447} & \textbf{21K} & 12.372 & 0.473 & \textbf{0.591} & 47K & \textbf{15.048} & \textbf{0.402} & \textbf{0.604} & \textbf{41K}
            \\  
            \toprule[0.5pt]
            % \specialrule{0.3pt}{0pt}{0pt}
            \multirow{2}{*}{SparseGS\cite{xiong2024sparsegs}}& -- & 13.599 & 0.325 & 0.513 & 268K & 16.781 & 0.599 & 0.374 & 499K & 16.414 & 0.430 & 0.464 & 600K\\
            &\textbf{+Ours} & \textbf{14.504} & \textbf{0.396} & \textbf{0.469} & \textbf{229K} & \textbf{17.333} & \textbf{0.626} & \textbf{0.364} & \textbf{483K} & \textbf{17.189} & \textbf{0.479} & \textbf{0.442} & \textbf{377K}
            \\
		\bottomrule[2pt]
	\end{tabular}
 \label{table:Sparse_Quantitative}
\end{table*}

\begin{figure*}[!t]  
	\centering
	\includegraphics[width=16.0cm]{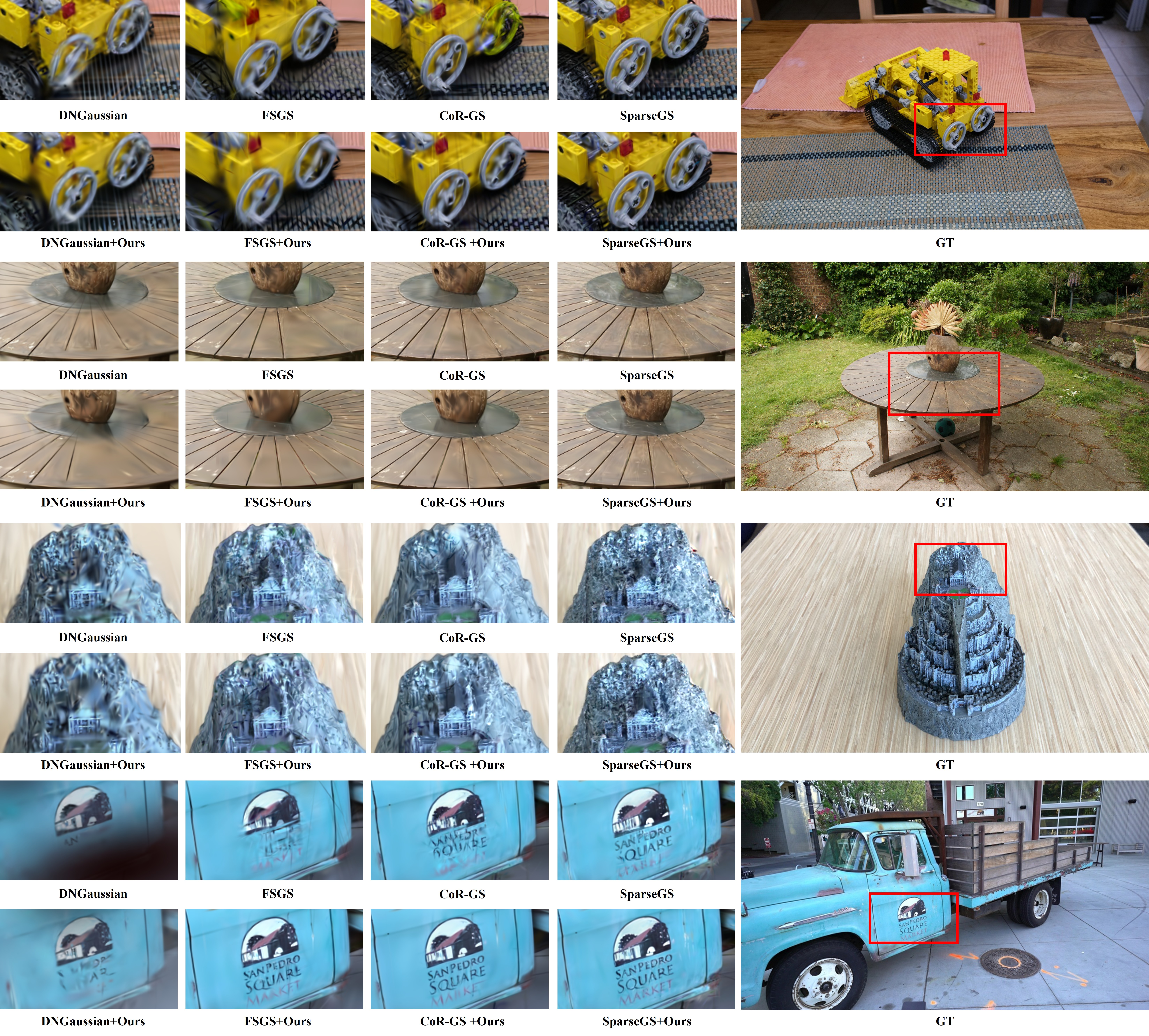}
    \vspace{-0.3cm}
	\caption{ Qualitative results before and after integrating our approach.
 }
	\label{Fig:spare_view}	
\end{figure*}

\noindent \textbf{Implementation Details.}
All experiments were conducted on an NVIDIA RTX A5000. 
During the implementation of ViewCrafter \cite{yu2024viewcrafter}, due to GPU memory limitations, we input 4 images at a time to generate validation images for the Mip-NeRF360 and Tanks datasets.
To perform joint initialization while avoiding the impact of generative hallucinations on camera intrinsics and extrinsics, we follow the strategy used in sparse-view settings. Specifically, we first estimate camera parameters using only the original input images. These parameters are then fixed, and the validation images are added to rerun COLMAP \cite{schoenberger2016sfm}. 
% This process provides camera poses for the validation views and yields an initial point cloud with better geometric structure.
For fair comparison across methods, the input point clouds used for FSGS and CoR-GS are extracted in the same way as most other methods, based on feature points.
During training, we apply validation monitoring every 100 iterations and perform Gaussian number optimization at 5k iterations. For DNGaussian, whose default training length is 6k iterations, we adjust the validation monitoring frequency to every 50 iterations and apply Gaussian optimization at 4k iterations.

\subsection{Comparison with Sparse-View Scenes}
As shown in Table \ref{table:Sparse_Quantitative}, DNGaussian performs poorly due to its reliance on scene-specific hyperparameters, which are not provided for the Tanks and Mip-NeRF360 datasets. In contrast, all other methods combined with our approach show significant performance improvements across three different datasets. Alongside better rendering quality, memory consumption is also greatly reduced.
For example, when combining our method with CoR-GS on the LLFF dataset, the PSNR on test views improves by 1.7dB, while the storage cost is reduced by 40$\%$. Figure \ref{Fig:spare_view_result} illustrates the results before and after applying our method to five different sparse-view 3DGS approaches on the Mip-NeRF360 dataset. With VGNC, all five methods demonstrate clear improvements in both rendering quality and memory efficiency. Qualitative comparisons in Figure \ref{Fig:spare_view} further show that our method significantly reduces the number of Gaussians while even enhancing details in some regions.

\subsection{Comparisons with Dense-View Scenes}
\begin{figure}[b]  
	\centering
	\includegraphics[width=8.2cm]{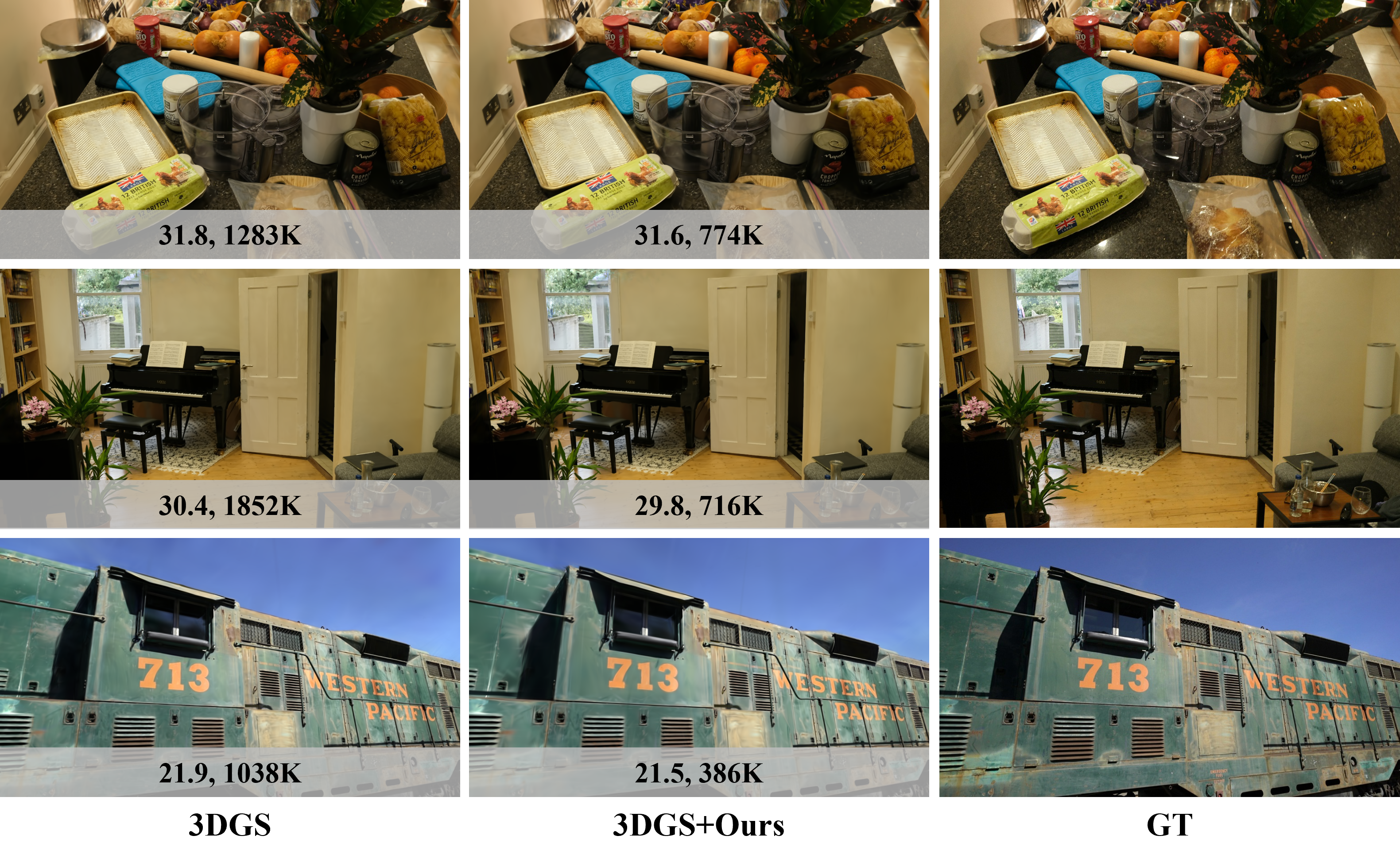}
    \vspace{-0.4cm}
	\caption{Qualitative comparisons in dense-view scenes. }
	\label{Fig:dense_view}	
\end{figure}

Our method is not only effective for mitigating overfitting in sparse-view 3DGS but can also be applied to dense-view reconstruction tasks. As shown in Table \ref{table:dens_360} and Table \ref{table:dens_tank}, our approach significantly reduces redundant Gaussians even in dense-view scenarios, while maintaining nearly the same rendering quality. Furthermore, the substantial reduction in Gaussian count leads to a notable acceleration in rendering speed. As demonstrated by the qualitative results in Figure \ref{Fig:dense_view}, our method enables 3DGS to preserve visual fidelity while substantially lowering the number of Gaussians.

\vspace{-0.2cm}

\begin{table}[!b]
	\caption{Dense-view setting of Mip-NeRF360 \cite{barron2022mip}.} 
    \vspace{-0.2cm}
	\centering
        \begin{tabular}{lccccc}  
	\toprule[1pt]
            Methods
        & PSNR$\uparrow$  & SSIM$\uparrow$ & LPIPS$\downarrow$ & Num.$\downarrow$ & FPS$\uparrow$ \\
        \toprule[0.5pt]
        3DGS\cite{kerbl20233d} & 28.117 & 0.843 & 0.172 & 3565K & 152\\
        3DGS + Ours & 27.884 & 0.832 & 0.194 & 1460K &  244\\
		\bottomrule[1pt]
	\end{tabular} 
	\label{table:dens_360}
\end{table}

\begin{table}[!b]
	\caption{Dense-view setting of tanks \cite{knapitsch2017tanks}.} 
    \vspace{-0.2cm}
	\centering
        \begin{tabular}{lccccc}  
	\toprule[1pt]
            Methods& PSNR$\uparrow$  & SSIM$\uparrow$ & LPIPS$\downarrow$ & Num.$\downarrow$ & FPS$\uparrow$\\
            \bottomrule[0.5pt]
        3DGS\cite{kerbl20233d} & 23.603 & 0.838 & 0.189 & 1751K & 136\\
        3DGS + Ours & 23.283 & 0.824 & 0.218 & 981K & 319 \\
		\bottomrule[1pt]
	\end{tabular} 
	\label{table:dens_tank}
\end{table}

\subsection{Ablation Studies}

\begin{figure}[t]  
	\centering
	%	\quad
	\subfloat[Sparse-view initialization]{
		\includegraphics[width=4.0cm]{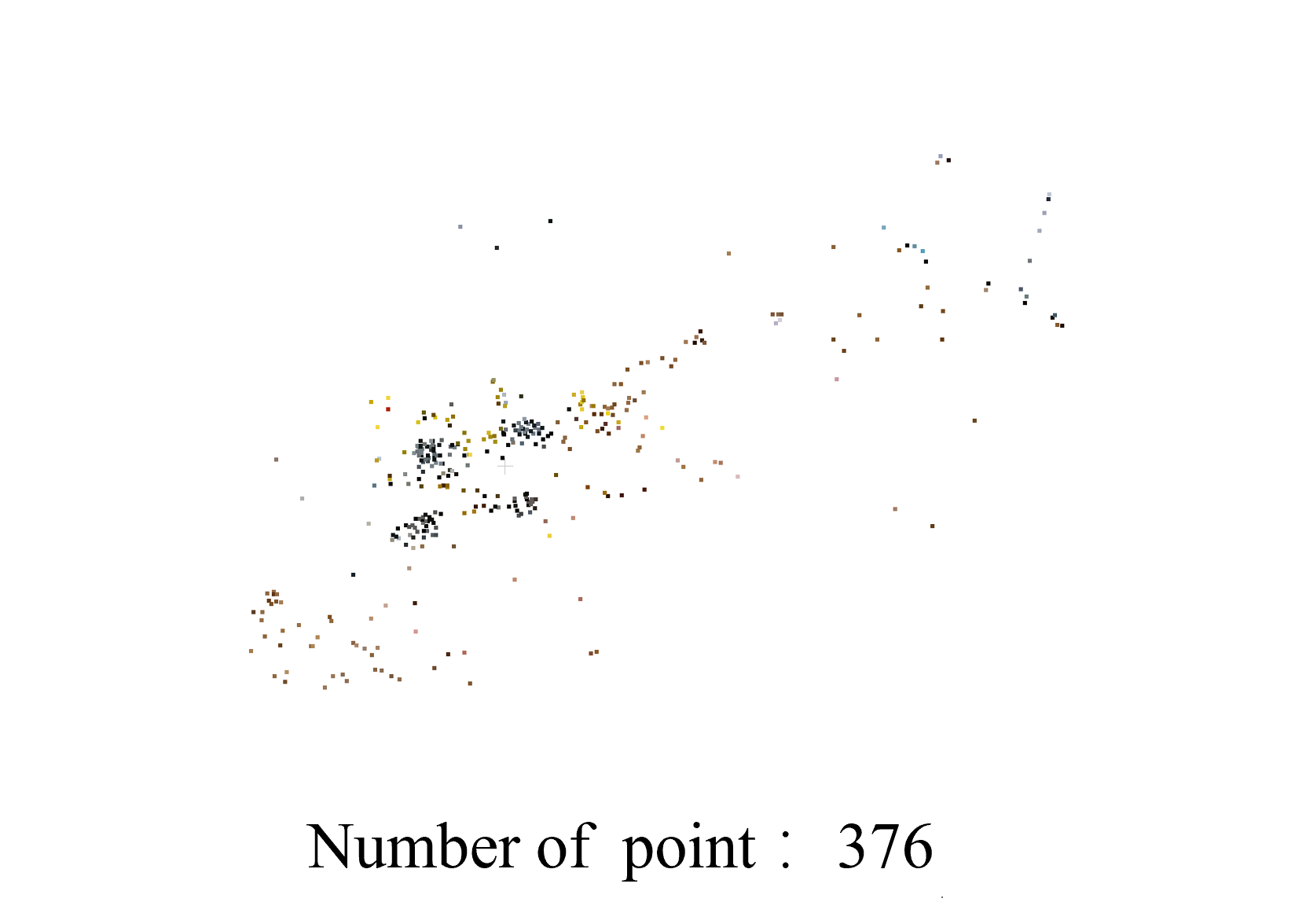}
	}
	\subfloat[Joint initialization]{
		\includegraphics[width=4.0cm]{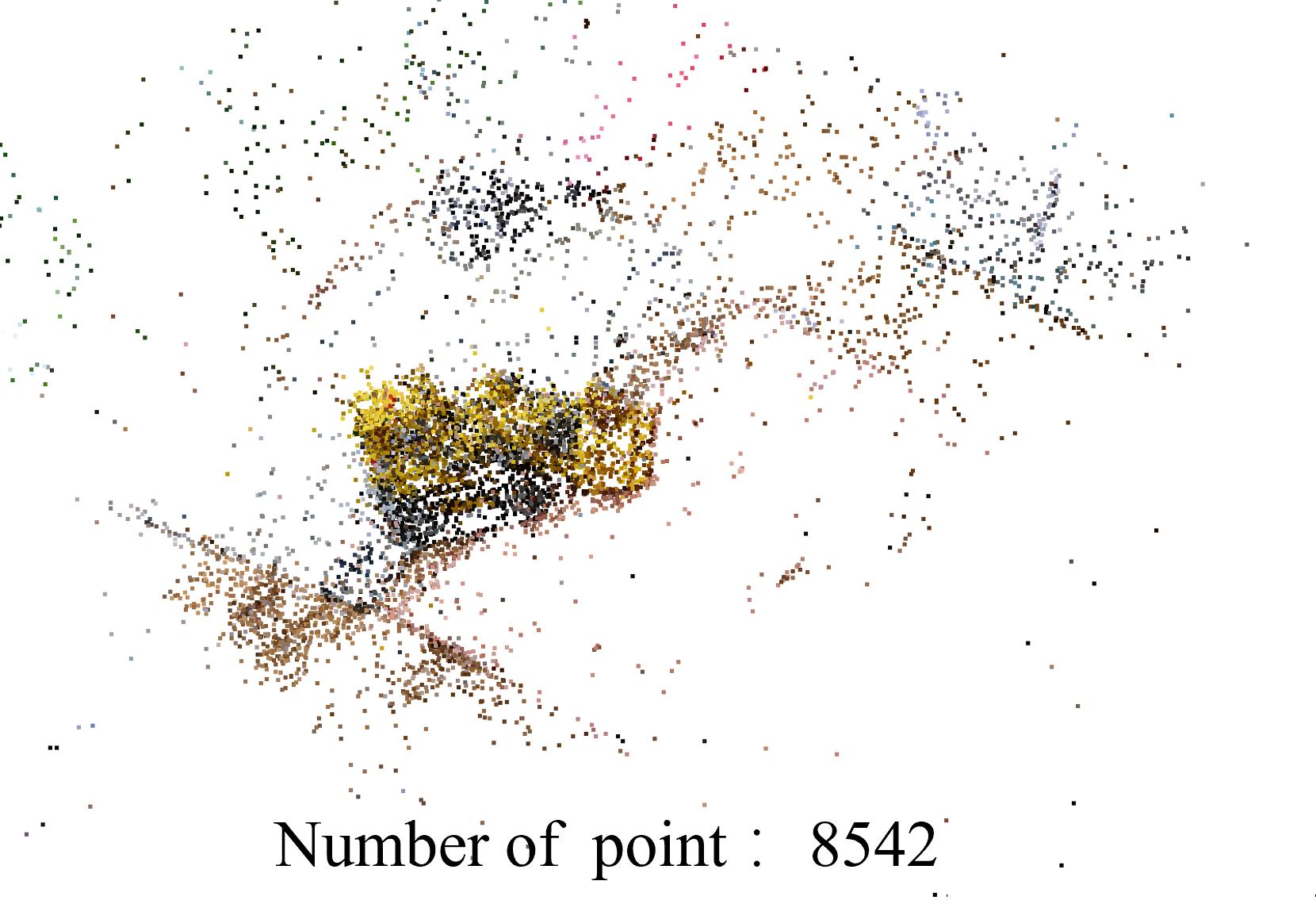}
	}
    \vspace{-0.4cm}
	\caption{Effectiveness of joint initialization.
 }
	\label{Fig:abla_joint}	
\end{figure}

We integrate each component of our method into SparseGS \cite{xiong2024sparsegs} on the Mip-NeRF360 dataset to validate their effectiveness. As shown in Figure \ref{Fig:abla_joint}, although the generated validation images contain some hallucinated regions, they also capture a wide range of correct 3D geometric information from novel viewpoints. Using these images for joint initialization helps generate a Gaussian point cloud with better geometry, which benefits the training process by placing the Gaussians in more appropriate positions.

As reported in Table \ref{table:ablation}, using validation images for initialization significantly improves rendering quality in sparse-view 3DGS. In addition, the validation-guided control of the Gaussian number alone can substantially reduce redundant points. This reduction not only shortens the rendering time but also speeds up training, while slightly improving rendering quality. Finally, combining both initialization and Gaussian number control leads to a significant decrease in the number of Gaussians and a notable improvement in novel-view rendering quality.

\begin{table}[!t]
	\caption{Ablation studies on SparseGS \cite{xiong2024sparsegs}} 
	\centering
    \resizebox{1.0\linewidth}{!}{
        \begin{tabular}{ccccccc}  
	\toprule[1pt]
    Joint ini.& Number Control& PSNR& SSIM& LPIPS& Num. & Train(min)\\
    \toprule[1pt]
    & & 16.247 & 0.424 & 0.468 & 601K & 32.0 \\
    \toprule[0.5pt]
    \checkmark & & \cellcolor{orange!30}17.077 & \cellcolor{orange!30}0.464 & \cellcolor{red!30}0.439 & 609K & 31.8 \\
    &\checkmark & 16.646 & 0.448 & 0.455 & \cellcolor{orange!30}499K & \cellcolor{red!30}25.9 \\
    \checkmark &\checkmark & \cellcolor{red!30}17.189 & \cellcolor{red!30}0.479 & \cellcolor{orange!30}0.446 & \cellcolor{red!30}377K & \cellcolor{orange!30}28.5 \\
		\bottomrule[1pt]
	\end{tabular} 
    }
	\label{table:ablation}
\end{table}

\section{Conclusion and Limitation}
We first reveal the overfitting issue in sparse-view  3DGS as the number of Gaussians increases. Then, we propose a validation image filtering method that incorporates generative NVS models to produce validation images for 3D reconstruction. Finally, we introduce VGNC, a validation-guided Gaussian Number Control strategy, which allows the model to automatically determine the optimal number of Gaussians during training, effectively reducing overfitting. More importantly, VGNC can be easily integrated into other sparse-view 3DGS frameworks to alleviate similar overfitting problems. Our experiments show that the proposed method can be applied to various sparse-view 3DGS approaches and effectively  improves reconstruction performance across diverse scenes.

Although our method makes effective use of generated images as validation data to guide Gaussian number control and initialization, some accurate geometric structures in these images remain underutilized. We hope this work inspires future research on better leveraging generative models—specifically, how to utilize reliable content from synthesized views while minimizing the impact of hallucinated artifacts.

\bibliographystyle{ACM-Reference-Format}
\bibliography{ValutationGaussian}

\makeatletter
\immediate\write18{bibtex \jobname} % 强制调用 BibTeX
\makeatother

%%
%% If your work has an appendix, this is the place to put it.
\appendix

\end{document}